\documentclass[lettersize,journal]{IEEEtran}
\usepackage{amsmath,amsfonts}
\usepackage{array}
\usepackage[caption=false,font=normalsize,labelfont=sf,textfont=sf]{subfig}
\usepackage{textcomp}
\usepackage{stfloats}
\usepackage{url}
\usepackage{verbatim}
\usepackage{graphicx}
\def\BibTeX{{\rm B\kern-.05em{\sc i\kern-.025em b}\kern-.08em
    T\kern-.1667em\lower.7ex\hbox{E}\kern-.125emX}}
\usepackage{balance}
\usepackage{graphicx}
\usepackage{lineno,hyperref}
\usepackage{amssymb, amsmath}
\usepackage{bm}
\usepackage{color}
\usepackage{threeparttable}
\usepackage{booktabs} 
\usepackage{graphicx}
\usepackage{cite}
\usepackage{color}
\usepackage{booktabs}
\usepackage{pifont}
\usepackage{array} 
\usepackage{multirow}
\usepackage{tabularx}
\usepackage{algorithm}
\usepackage{algpseudocode}
\usepackage{tabularray}
\usepackage{makecell}
\usepackage{xcolor} 
\usepackage{booktabs} 

\algrenewcommand\algorithmicrequire{\textbf{Input:}}
\algrenewcommand\algorithmicensure{\textbf{Output:}}

\makeatletter
\newcommand{\removelatexerror}{\let\@latex@error\@gobble}
\makeatother

\begin{document}

\title{CroBIM-U: Uncertainty-Driven Referring Remote Sensing Image Segmentation}

\author{
	Yuzhe~Sun,
	Zhe~Dong,
	Haochen~Jiang,
	Tianzhu~Liu,~\IEEEmembership{Member,~IEEE},
	and~Yanfeng~Gu,~\IEEEmembership{Senior Member,~IEEE}
	
\thanks{Manuscript received XX xx, 2026; revised XX xx, 2026; accepted XX xx, 2026.}
\thanks{This work was supported by the National Natural Science Foundation of China under Grant 624B2051. 
\emph{(Corresponding author: Yanfeng Gu.)}}
\thanks{Y. Sun, Z. Dong, H. Jiang, T. Liu and Y. Gu are with the School of Electronics and Information Engineering, Harbin Institute of Technology, Harbin 150001, China. (email: guyf@hit.edu.cn).}
}

\maketitle

\begin{abstract}
	Referring remote sensing image segmentation aims to localize specific targets described by natural language within complex overhead imagery. However, due to extreme scale variations, dense similar distractors, and intricate boundary structures, the reliability of cross-modal alignment exhibits significant \textbf{spatial non-uniformity}. Existing methods typically employ uniform fusion and refinement strategies across the entire image, which often introduces unnecessary linguistic perturbations in visually clear regions while failing to provide sufficient disambiguation in confused areas. To address this, we propose an \textbf{uncertainty-guided framework} that explicitly leverages a pixel-wise \textbf{referring uncertainty map} as a spatial prior to orchestrate adaptive inference. Specifically, we introduce a plug-and-play \textbf{Referring Uncertainty Scorer (RUS)}, which is trained via an online error-consistency supervision strategy to interpretably predict the spatial distribution of referential ambiguity. Building on this prior, we design two plug-and-play modules: 1) \textbf{Uncertainty-Gated Fusion (UGF)}, which dynamically modulates language injection strength to enhance constraints in high-uncertainty regions while suppressing noise in low-uncertainty ones; and 2) \textbf{Uncertainty-Driven Local Refinement (UDLR)}, which utilizes uncertainty-derived soft masks to focus refinement on error-prone boundaries and fine details. Extensive experiments demonstrate that our method functions as a unified, plug-and-play solution that significantly improves robustness and geometric fidelity in complex remote sensing scenes without altering the backbone architecture.
\end{abstract}

\begin{IEEEkeywords}
	Referring remote sensing image segmentation, uncertainty estimation, spatial non-uniformity, cross-modal fusion, plug-and-play module.
\end{IEEEkeywords}

\section{Introduction}

Remote-sensing imagery, characterized by broad spatial coverage and frequent revisits, is fundamental to applications such as urban mapping, disaster assessment, ecological monitoring, and resource inventory \cite{yuan2023rrsis, zhan2024rsvg, dong2023distilling}. In many real-world scenarios, however, users require query flexibility beyond pre-defined semantic categories (e.g., \emph{building} or \emph{road}). Instead, they often specify targets via natural-language descriptions of attributes, spatial relations, and context (e.g., \emph{the slender bridge crossing the river next to the port}). This demand has motivated the emergence of \emph{referring remote sensing image segmentation} (RRSIS) \cite{yuan2023rrsis, liu2024rotated}: given an image and a text description, the model predicts a pixel-wise mask of the referred instance. Compared with category-driven semantic segmentation, RRSIS enables more flexible human--computer interaction and supports instance-level, fine-grained mapping over large-scale geospatial data.

Compared with referring segmentation in natural scenes (RIS) \cite{yang2022lavt}, remote-sensing settings pose severe and unique challenges in cross-modal localization and boundary delineation \cite{yuan2023rrsis, rsrefseg2025}. First, remote-sensing scenes are often dense and repetitive; visually similar instances (e.g., rooftops, ships, road fragments, and farmland parcels) can easily cause referential ambiguity \cite{liu2024rotated}, making cross-modal alignment reliability highly spatially non-uniform. Second, targets vary drastically in scale and orientation \cite{liu2024rotated, yuan2023rrsis}, and the overhead viewpoint weakens viewpoint-dependent cues (e.g., perspective), increasing confusion under compositional descriptions. Third, many targets exhibit slender, fragmented, or low-contrast structures (e.g., roads, coastlines, dikes, and building contours), where errors tend to concentrate around boundaries \cite{refinedunet2020}. These observations suggest that, even within a single image, both \emph{referential reliability} (whether the correct instance is localized) and \emph{pixel prediction reliability} (whether local geometry is stable) can vary substantially across space. Consequently, applying uniform fusion strength and refinement computation across all regions---as is common in current practices---may under-constrain high-risk locations while unnecessarily perturbing already reliable ones.

Existing approaches typically follow a two-tower design that encodes the image and text separately, predicting masks through cross-modal interaction in a decoder \cite{yang2022lavt, yuan2023rrsis}. While recent works have introduced multi-scale interactions \cite{liu2024rotated}, foundation model priors (e.g., CLIP-SAM based methods) \cite{rsrefseg2025, dong2024upetu}, or even generative diffusion models \cite{dong2025diffris} to handle scale and semantic gaps, many implementations still lack an explicit, error-aligned spatial prior. Furthermore, although advanced generative foundation models \cite{dong2024generative} have improved visual representation, efficiently adapting them to pixel-level referring tasks with limited spatial selectivity remains challenging. In RRSIS, a uniform fusion strategy can induce two coupled issues. In visually confident regions with stable predictions, overly aggressive language modulation may introduce unnecessary perturbations. Conversely, in high-risk regions with severe ambiguity or fragile geometry (especially boundaries and thin structures), uniform fusion capacity can be insufficient for disambiguation and correction. This motivates \emph{risk-aware} modeling: beyond predicting \emph{what} to segment, the model should estimate \emph{where} mistakes are more likely and allocate cross-modal constraints and refinement resources accordingly.

To this end, we introduce a pixel-wise cross-modal referential uncertainty map as an intermediate representation of spatial risk. Here, uncertainty serves as a risk cue indicating that a location is more likely to suffer referential confusion or local segmentation error. During inference, the uncertainty map acts as a feed-forward spatial prior to enable selective cross-modal fusion and focused local correction: language constraints are strengthened at high-uncertainty locations to resolve ambiguity, while unnecessary cross-modal perturbations are suppressed at low-uncertainty locations to preserve stable predictions. Moreover, the uncertainty prior concentrates refinement on error-prone areas such as boundaries and thin structures, improving geometric fidelity and mask consistency.

A key requirement is that the predicted uncertainty should be interpretable and semantically aligned with actual errors. Network design alone does not guarantee that uncertainty correlates with mistakes, especially under the complex ambiguities of remote-sensing scenes. We therefore propose an online risk learning strategy: during training, we derive a per-pixel error signal from the discrepancy between the model's current prediction and the ground truth, and supervise uncertainty to be positively correlated with this error signal. This online, behavior-adaptive supervision unifies cross-modal disambiguation difficulty and boundary/detail instability into a single spatial prior that consistently guides fusion control and local correction.

Building on the uncertainty prior, we develop a risk-guided fusion--local correction pipeline with three plug-and-play modules. First, the Referring Uncertainty Scorer (RUS) predicts the pixel-wise uncertainty map on top of encoder features and learns risk semantics via the proposed online error supervision. Second, Uncertainty-Gated Fusion (UGF) is a lightweight gated fusion block that modulates language injection strength conditioned on uncertainty, strengthening cross-modal constraints in uncertain regions while suppressing ineffective perturbations in confident ones. Third, Uncertainty-Driven Local Refinement (UDLR) generates shape-adaptive soft refinement masks from uncertainty and guides a lightweight residual branch to focus corrections on high-risk regions (e.g., boundaries and thin structures), reducing merging/bleeding, discontinuities, and boundary shifts. Overall, without modifying the two-tower backbone, our method provides a unified mechanism for online risk learning during training and risk-guided fusion and correction during inference, matching the pronounced spatial non-uniformity of reliability in remote-sensing referring segmentation.

Our contributions are summarized as follows: \begin{itemize} \item We propose a risk-aware framework for remote-sensing referring segmentation, introducing a pixel-wise cross-modal referential uncertainty map as a unified representation of spatial risk to guide inference. \item We design RUS and an online error-aligned supervision strategy to make uncertainty prediction semantically consistent with model errors, yielding an interpretable spatial risk prior. \item We propose two plug-and-play modules, UGF and UDLR, to adaptively control language modulation and to perform shape-adaptive local refinement driven by uncertainty, improving disambiguation and boundary/detail quality in high-risk regions while preserving stability in low-risk regions. \item We build a risk-guided fusion--local correction pipeline that enhances robustness and geometric fidelity in complex remote-sensing scenes without changing the backbone architecture. \end{itemize}

\section{Related Work}

\subsection{Referring Image Segmentation in Natural Scenes}
Referring Image Segmentation (RIS) requires fine-grained semantic alignment between visual and linguistic modalities. The field has evolved from CNN-RNN architectures to Transformer-based and generative paradigms.

\subsubsection{Evolution of Architectures}
Early methods adopted a ``concat-and-process'' pipeline. Hu \textit{et al.} \cite{hu2016segmentation} pioneered the task using CNNs and LSTMs with simple concatenation. To enhance cross-modal interaction, attention mechanisms became standard. Shi \textit{et al.} \cite{shi2018key} focused on linguistic key-words, while Ye \textit{et al.} \cite{ye2019cross} employed cross-modal self-attention. Recognizing the complexity of language, subsequent works explored progressive fusion strategies. Feng \textit{et al.} \cite{feng2021encoder} introduced an encoder fusion network to interleave features early, and Huang \textit{et al.} \cite{huang2020referring} proposed a progressive comprehension framework to refine segmentation iteratively. Furthermore, multi-task learning frameworks \cite{luo2020multi} have been explored to jointly optimize referring expression comprehension and segmentation.

\subsubsection{Transformer and Generative Models}
The introduction of Transformers \cite{vaswani2017attention} revolutionized RIS. VLT \cite{ding2021vision} and LAVT \cite{yang2022lavt} leverage hierarchical vision transformers to perform early vision-language fusion, suppressing background noise effectively. Recent trends include foundation model adaptation, such as CRIS \cite{wang2022cris} using CLIP \cite{radford2021learning}, and generative approaches like SeqTR \cite{zhu2022seqtr} and PolyFormer \cite{liu2023polyformer} that frame segmentation as sequence prediction. Additionally, efforts have been made to enhance robustness against adversarial or ambiguous inputs \cite{wu2022towards}, and to model temporal-spatial context in dynamic scenes \cite{tang2023referring}.

Despite these successes in object-centric natural images, direct transfer to remote sensing is hindered by the domain gap, specifically the lack of canonical viewpoints and the prevalence of dense, tiny objects.

\subsection{Referring Remote Sensing Image Segmentation}
RS-RefSeg extends grounding to overhead imagery, addressing challenges like extreme scale variation and rotation. The release of datasets such as GeoRef \cite{yuan2022language}, RefSeg-RS \cite{zhan2024rrsis}, and the large-scale SkyEye \cite{guo2023skyeye} has established the benchmark for this field.

\subsubsection{Scale and Rotation Invariance}
Initial works focused on geometric adaptations. Yuan \textit{et al.} \cite{yuan2022language} addressed scale disparity via global-local attention. To handle arbitrary orientations, Liu \textit{et al.} \cite{liu2023rotation} proposed a rotation-invariant transformer. Recent works continue to refine multi-scale reasoning; for instance, Zheng \textit{et al.} \cite{zheng2024mmrs} developed a multi-modal multi-scale fusion network to capture objects at varying resolutions, while Xu \textit{et al.} \cite{xu2024glrs} emphasized global-local reasoning to contextualize targets within broad geographic scenes.

\begin{figure*}[tbp]
	\begin{center}
		\centerline{\includegraphics[width=1\linewidth]{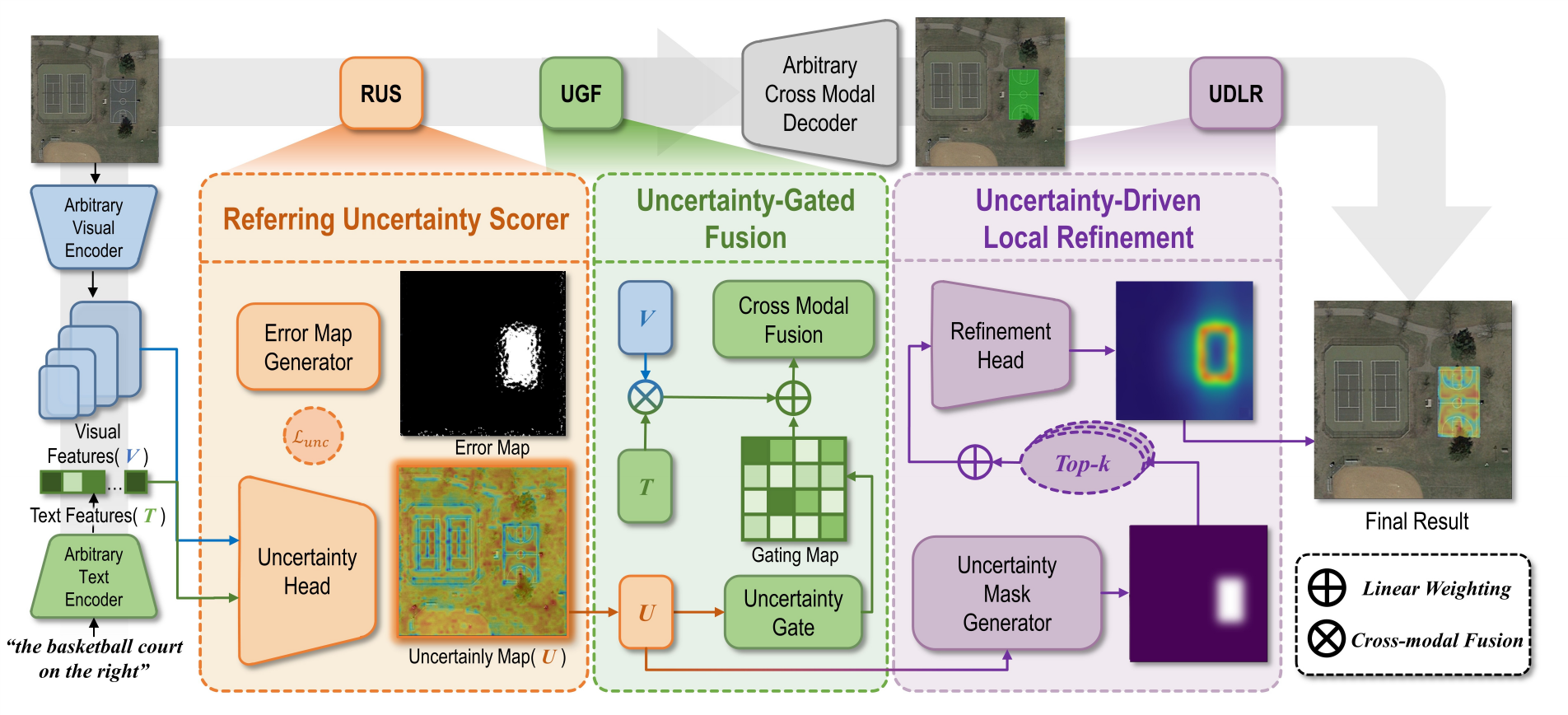}}
		\caption{Flowchart of the proposed risk-aware remote-sensing referring segmentation framework. Given an image $I$ and a referring expression $S$, an arbitrary visual encoder and text encoder extract visual features $\mathbf{V}$ and text features $\mathbf{T}$. The Referring Uncertainty Scorer (RUS) predicts a pixel-level grounding uncertainty map $U$ (logits), which is supervised during training by an online error map generated from the decoder prediction, yielding the uncertainty loss $\mathcal{L}_{\mathrm{unc}}$ (the error-map branch is used only in training). The predicted $U$ provides a spatial prior to (i) Uncertainty-Gated Fusion (UGF), which performs a one-shot uncertainty-conditioned post-fusion to modulate language injection and produce fused features for an arbitrary cross-modal decoder, and (ii) Uncertainty-Driven Local Refinement (UDLR), which converts $U$ into a soft refinement mask to guide a lightweight refinement head for boundary and fine-structure correction, producing the final segmentation. $\oplus$ denotes linear weighting and $\otimes$ denotes cross-modal fusion.}\label{flowchart}
	\end{center}
\end{figure*}

\subsubsection{Cross-Modal Alignment and Refinement}
Distinguishing targets from similar distractors in dense clusters remains difficult. Zhang \textit{et al.} \cite{zhang2023multi} utilized multi-granularity interaction, and Ma \textit{et al.} \cite{ma2024referring} proposed cross-modal progressive fusion to align features step-by-step. To improve boundary precision, Sun \textit{et al.} \cite{sun2024boundary} introduced boundary-aware attention. Several works focus on consistency and alignment; Wang \textit{et al.} \cite{wang2023bilateral} designed a bilateral alignment network, and Ji \textit{et al.} \cite{ji2023advancing} enforced spatial-linguistic consistency to reduce grounding errors. Adapting Large Vision-Language Models (LVLMs) is also a growing trend, with works like RS-Prompter \cite{chen2024rsprompter} and GeoRSCLIP \cite{li2024georsclip} demonstrating strong generalization \cite{zhu2024visual, fu2023language}.

\textbf{Critique:} While existing methods excel at feature extraction and alignment, they typically apply a uniform inference mechanism across the image. They lack \emph{spatial risk awareness}, treating ambiguous regions (e.g., shadows, crowded parking lots) with the same confidence as clear backgrounds. Our work postulates that explicitly modeling this uncertainty is key to breaking the current performance ceiling.

\subsection{Cross-Modal Uncertainty Estimation}
Uncertainty quantification \cite{abdar2021review} distinguishes between aleatoric (data) and epistemic (model) uncertainty \cite{kendall2017uncertainties}.

\subsubsection{Uncertainty in Vision}
In standard segmentation, Bayesian approximations like MC Dropout \cite{gal2016dropout} and Deep Ensembles \cite{lakshminarayanan2017simple} are classic but costly. Related Bayesian formulations have also been explored in cross-modal settings such as VQA question generation \cite{bhattacharya2019bayesian}. Deterministic methods \cite{van2020uncertainty, sensoy2018evidential} offer efficiency. Generative models, such as the Probabilistic U-Net \cite{kohl2018probabilistic}, model the distribution of plausible segmentations for ambiguous images. These concepts have been applied to 3D detection \cite{yu2019uncertainty}, zero-shot learning \cite{hu2020uncertainty}, and medical imaging \cite{nair2020exploring, judge2022uncertainty}, providing general frameworks for reliability \cite{loquercio2020general}.

\subsubsection{Uncertainty in Remote Sensing}
In RS, uncertainty has primarily been used as a passive indicator for post-processing in land cover classification \cite{kampffmeyer2016semantic}, change detection \cite{li2022uncertainty}, or active learning \cite{wang2021active}. Boundary-aware graph reasoning \cite{li2021boundary} and context-aware networks \cite{fan2021context} implicitly handle ambiguity but do not explicitly quantify it.

\textbf{Our Contribution:} Unlike prior works that use uncertainty merely for quality assessment, we propose an \emph{active uncertainty-guided framework}. We leverage the cross-modal uncertainty map as a dynamic weight to orchestrate feature refinement, forcing the model to re-attend to high-risk regions where visual-linguistic alignment is fragile.

\section{Methodology}

\subsection{Overview}

We study remote-sensing referring segmentation: given a remote-sensing image $I$ and a text description $S$, the goal is to predict the pixel-wise mask of the referred target $\hat{Y}\in[0,1]^{H\times W}$ \cite{yuan2023rrsis}. A central challenge is the spatially varying reliability of both cross-modal grounding and pixel prediction. On the one hand, abundant visually similar instances and large scale variations in remote-sensing imagery induce ambiguities in vision--language alignment \cite{liu2024rotated, zhan2024rsvg}; on the other hand, thin or fragmented structures and blurred boundaries cause errors to cluster around object boundaries and fine details. As a result, uniformly injecting language cues or allocating uniform refinement across the entire image often leads to two coupled issues: unnecessary language-induced interference in reliable regions, and insufficient disambiguation as well as inadequate compute allocation in high-risk regions.

To explicitly capture such a spatial risk distribution, we introduce a pixel-level cross-modal grounding uncertainty map $U$ as the key intermediate representation. Unless otherwise specified, $U$ is represented in \emph{logit} form; when a probabilistic uncertainty is needed, we use $U^{p}=\sigma(U)\in[0,1]$, where $\sigma(\cdot)$ denotes the Sigmoid function. By learning $U$ with online supervision during training, the model predicts where errors are more likely to occur. At inference time, $U$ serves as a feed-forward spatial prior for selective fusion and focused local refinement, forming a risk-aware pipeline tailored to remote-sensing referring segmentation.

Most existing methods adopt a two-tower architecture \cite{yang2022lavt, ding2021vision}: a visual encoder and a text encoder extract representations separately, followed by a pixel-level decoder that performs cross-modal fusion and produces the segmentation output. Concretely, the visual encoder extracts multi-scale visual features while the text encoder maps $S$ to token embeddings. During progressive decoding, cross-modal interactions produce a foreground logit map $P_{\mathrm{fg}}\in\mathbb{R}^{B\times 1\times H\times W}$ and an initial mask prediction $\hat{Y}=\sigma(P_{\mathrm{fg}})$.

Built upon this backbone, we introduce three plug-and-play modules that share a unified intermediate representation---the grounding uncertainty map $U$. (1) \textbf{Referring Uncertainty Scorer (RUS)} is placed between the encoder and decoder. It takes visual and textual tokens as inputs and predicts $U$ to quantify spatial grounding ambiguity and local error risk. (2) \textbf{Uncertainty-Gated Fusion (UGF)} is applied \emph{once} as a post-fusion step before decoding: it modulates the strength of language injection according to $U$, so that textual constraints primarily affect high-uncertainty locations while suppressing interference in low-uncertainty regions. (3) \textbf{Uncertainty-Driven Local Refinement (UDLR)} derives shape-adaptive soft regions from $U^{p}$ to guide a refinement branch, concentrating corrections on error-prone areas such as boundaries and fine structures. Together, these components instantiate a training-time ``online risk learning'' loop (via RUS supervision) and an inference-time ``risk-guided fusion--local correction'' pipeline.

\subsection{Backbone and Interaction Stages}
\label{section:backbone}

Following two-tower referring segmentation pipelines \cite{yang2022lavt}, we instantiate the framework with a ConvNeXt-B visual encoder \cite{liu2022convnet}, a BERT text encoder \cite{devlin2018bert}, and a Mask2Former-style multi-scale mask decoder \cite{cheng2022masked}.
The visual encoder outputs a four-level feature pyramid at strides $\{4,8,16,32\}$ with channel dimensions $\{256,512,1024,2048\}$.
Each level is projected to a unified embedding dimension ($C{=}256$) and flattened into visual tokens
$\mathbf{V}^{(s)} \in \mathbb{R}^{B \times N_s \times C}$ with $N_s = H_s W_s$.
The text encoder produces token embeddings which are projected to the same dimension, yielding
$\mathbf{T}\in \mathbb{R}^{B\times L\times C}$ with a padding mask $\mathbf{m}\in\{0,1\}^{B\times L}$ (where $\mathbf{m}=1$ indicates the token is ignored).

The decoder follows a Mask2Former-style design \cite{cheng2022masked} with transformer-based multi-scale interaction.
In our instantiation, the multi-scale transformer encoder contains $6$ layers with $8$ attention heads and operates on three feature levels (strides $\{8,16,32\}$).
Let $\Phi^{(s)}(\cdot)$ denote one interaction block that updates visual tokens by cross-modal interaction and subsequent token mixing. At stage $s$, we have
\begin{equation}
	\begin{aligned}
		\mathbf{V}^{(s)+} &= \Phi^{(s)}\big(\mathbf{V}^{(s)}, \mathbf{T}, \mathbf{m}\big), \\
		\mathbf{F}^{(s)+} &= \mathrm{reshape}\big(\mathbf{V}^{(s)+}\big),
	\end{aligned}
\end{equation}
where $\mathrm{reshape}(\cdot)$ restores the spatial layout $(H_s, W_s)$.
The decoder finally outputs foreground logits $P_{\mathrm{fg}}\in\mathbb{R}^{B\times 1\times H\times W}$, producing $\hat{Y}=\sigma(P_{\mathrm{fg}})$.

RUS predicts an uncertainty logit map $U\in\mathbb{R}^{B\times 1\times H_u\times W_u}$ at token resolution. In our design, RUS operates on the \emph{coarsest} (stride-$32$) visual tokens, which have the largest receptive field; thus $(H_u,W_u)=(H_{32},W_{32})$.
When used at a decoder stage $s$ or at full resolution, we align $U$ by bilinear interpolation \emph{on logits}:
\begin{equation}
	\begin{aligned}
		U^{(s)} &= \mathrm{Resize}\big(U, H_s, W_s\big), \\
		\tilde{U} &= \mathrm{Resize}\big(U, H, W\big).
	\end{aligned}
\end{equation}
Interpolating in logit space avoids probability saturation and provides a consistent spatial prior for uncertainty-gated fusion and local refinement.

\begin{figure}[tbp]
	\begin{center}
		\centerline{\includegraphics[width=1\linewidth]{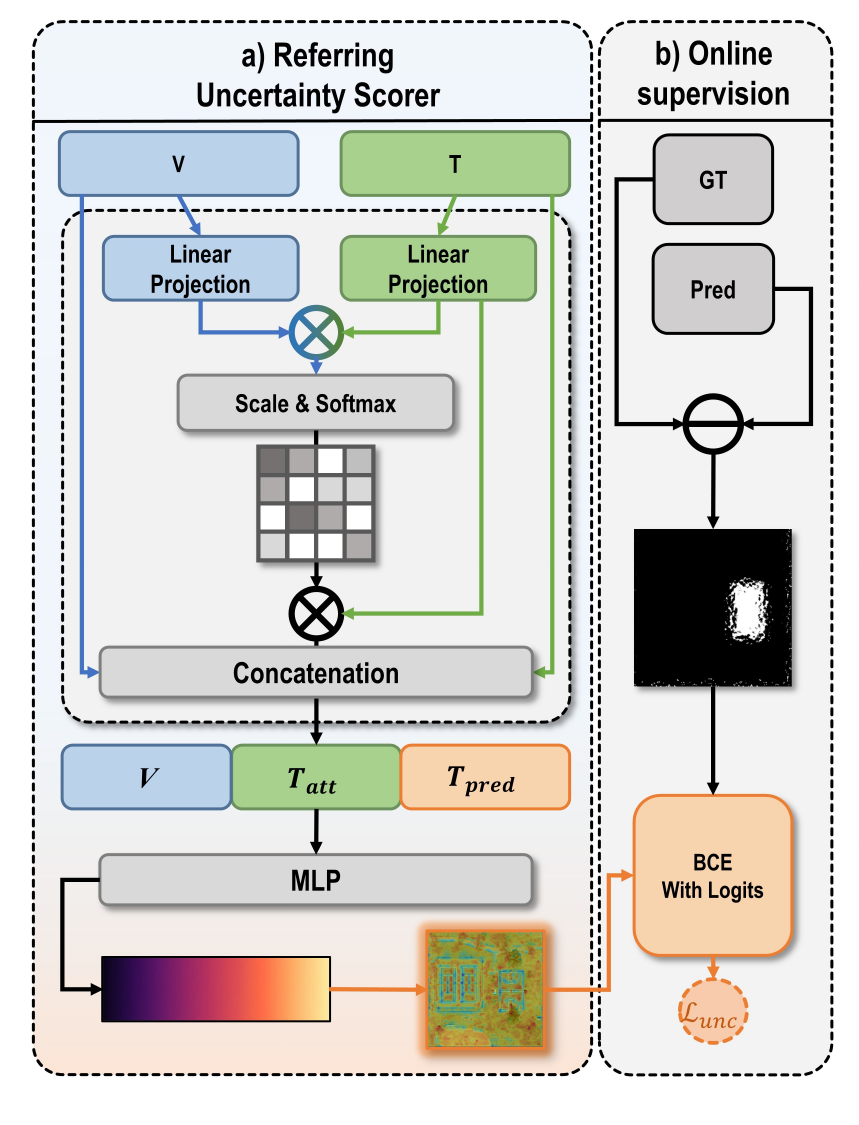}}
		\caption{Overview of Referring Uncertainty Scorer (RUS). \textbf{a)} Fuses visual ($V$) and textual ($T$) features via cross-attention and raw broadcasting to predict uncertainty maps. \textbf{b)} Trains with an error-consistency loss ($\mathcal{L}_{unc}$) derived from the prediction-GT discrepancy ($\ominus$).}\label{RUS}
	\end{center}
\end{figure}

\subsection{Referring Uncertainty Scorer}
\label{sec:rus}

Errors in remote-sensing referring segmentation are often spatially clustered: semantically similar land-cover types induce confusion, while thin/fractured structures and blurred boundaries cause locally unstable predictions \cite{yuan2023rrsis}. To explicitly model this location-dependent risk distribution \emph{before} decoding and multi-scale fusion, we insert a Referring Uncertainty Scorer (RUS) between the encoder and decoder to predict a pixel-wise referring uncertainty map $U$.

\paragraph{Architecture}
Let $\mathbf{V} \in \mathbb{R}^{B \times N_u \times C}$ denote the stride-$32$ visual tokens (after projection to $C{=}256$), and $\mathbf{T}\in\mathbb{R}^{B\times L\times C}$ denote the projected text tokens.
RUS outputs an uncertainty logit $\mathbf{u} \in \mathbb{R}^{B \times N_u \times 1}$ for each visual position, and reshapes it into
$U \in \mathbb{R}^{B \times 1 \times H_u \times W_u}$.
We define $U$ as an indicator of \emph{text-referential matching ambiguity / local error risk}: a larger value suggests that the position is more likely to suffer from confusion among similar objects or boundary/detail mis-segmentation, and subsequent modules should enforce stronger textual constraints or allocate more refinement capacity there.

To extract position-level referential ambiguity cues with minimal overhead, we apply linear projections to the visual and textual features (initialized as identity/near-identity after projection for stable early training) and employ scaled dot-product attention to perform ``visual-position to text-token'' alignment and aggregation. We then concatenate the original visual token with the aligned textual feature and feed them into a lightweight MLP to output the uncertainty logit:
\begin{equation}
	\begin{aligned}
		\mathbf{V}' &= \mathbf{V}\mathbf{W}_v, \quad \mathbf{T}' = \mathbf{T}\mathbf{W}_t, \\
		\mathbf{A} &= \mathrm{Softmax}\left(\gamma \cdot \mathbf{V}' {\mathbf{T}'}^{\top}\right), \\
		\tilde{\mathbf{T}} &= \mathbf{A}\mathbf{T}', \\
		\mathbf{u} &= \mathrm{MLP}\left( \left[ \mathbf{V} ; \tilde{\mathbf{T}} \right] \right).
	\end{aligned}
\end{equation}
Here, $\mathrm{Softmax}(\cdot)$ is applied along the text-sequence dimension, $\gamma$ is a learnable scaling factor initialized to the standard attention scale (e.g., $1/\sqrt{C}$), and $\mathbf{W}_v,\mathbf{W}_t$ are linear projection matrices. Finally, $U=\mathrm{reshape}(\mathbf{u})$ recovers the spatial layout $(H_u,W_u)$. The RUS output is kept in \emph{logit} form (without Sigmoid), matching the subsequent gating mechanism and the training-time supervision for numerical consistency.

\paragraph{Online supervision for risk semantics}
Architecture alone does not guarantee that $U$ learns the desired ``risk hint'' semantics; therefore, we construct an online pixel-wise supervision signal during training. Let the decoder output foreground logits $P_{\mathrm{fg}} \in \mathbb{R}^{B \times 1 \times H \times W}$. We obtain a binary prediction
\begin{equation}
	\hat{y} \;=\; \mathrm{StopGrad}\!\left(\mathbb{I}\big(\sigma(P_{\mathrm{fg}}) > 0.5\big)\right),
\end{equation}
and build a pixel-wise error map
\begin{equation}
	e(x) = \mathbb{I}\big(\hat{y}(x) \neq y(x)\big),
\end{equation}
where $y(x)$ is the ground truth (GT) and $\mathrm{StopGrad}(\cdot)$ prevents gradients from backpropagating through the online target.
We \emph{fix} the supervision target as $z(x)=e(x)$, so that larger $U$ consistently corresponds to higher error risk.
Optionally, we apply a light local smoothing (e.g., small-kernel blur) to suppress isolated noisy pixels, yielding $z(x)=\mathrm{Blur}(e(x))$.

Since the resolution of $U$ is lower than that of the final segmentation, we align it to the segmentation resolution to obtain $\tilde{U} \in \mathbb{R}^{B \times 1 \times H \times W}$ and optimize a weighted per-pixel BCEWithLogits loss:
\begin{equation}
	\label{eq:l_unc}
	\begin{aligned}
		\mathcal{L}_{\text{unc}} = &\frac{1}{BHW} \sum_{b,x} w_b^{\text{sample}} \cdot w_{b,x}^{\text{pixel}} \\
		&\cdot \mathrm{BCEWithLogits}(\tilde{U}_b(x), z_b(x))
	\end{aligned}
\end{equation}

We use two lightweight reweighting terms for stable online risk learning. (i) Sample-level difficulty weight:
\begin{equation}
	w_b^{\text{sample}} = 1 + \lambda_s \cdot \mathbb{I}\big(\mathrm{IoU}(\hat{y}_b, y_b) < \delta\big),
\end{equation}
which upweights globally hard samples (we use $\delta{=}0.5$ and $\lambda_s{=}1$ unless otherwise stated).
(ii) Pixel-level class rebalancing weight to counter sparsity of error pixels:
\begin{equation}
	w_{b,x}^{\text{pixel}} =
	\begin{cases}
		\frac{n_{b}^{0}+\epsilon}{n_{b}^{1}+\epsilon} & \text{if } z_b(x)=1, \\
		1 & \text{if } z_b(x)=0,
	\end{cases}
\end{equation}
where $n_b^{1}$ and $n_b^{0}$ denote the counts of positive/negative pixels in $z_b$, and $\epsilon$ is a small constant.
To mitigate early-stage noise in the online supervision, we enable $\mathcal{L}_{\mathrm{unc}}$ only after a short warm-up period.

Finally, $\mathcal{L}_{\mathrm{unc}}$ complements the segmentation loss $\mathcal{L}_{\mathrm{seg}}$:
$\mathcal{L}_{\mathrm{seg}}$ learns \emph{what} to segment, while $\mathcal{L}_{\mathrm{unc}}$ learns \emph{where} referring/segmentation errors are more likely.
Through online supervision, RUS explicitly encodes this distribution into $U$, providing consistent spatial guidance for uncertainty-gated fusion and soft-region local refinement.

%
%
%

\subsection{Uncertainty-Gated Fusion}
\label{sec:ugf}

In remote-sensing referring segmentation, linguistic cues exhibit a dual effect of disambiguation and perturbation. In regions crowded with visually similar objects, textual constraints are crucial for distinguishing the target; however, uniformly injecting language information even where visual evidence is already sufficient can induce referring drift and amplify mis-segmentation \cite{liu2024rotated}. To address this spatially non-uniform reliability, we propose Uncertainty-Gated Fusion (UGF), which uses the pixel-wise referring uncertainty predicted by RUS to modulate cross-modal updates at each position, enabling on-demand language allocation: strengthening disambiguation in high-risk regions while suppressing ineffective perturbations in low-risk regions.

UGF is instantiated as a \emph{single post-fusion} module applied between the encoder and the decoder (i.e., not repeated across decoding stages). Let $\mathbf{V} \in \mathbb{R}^{B \times N \times C}$ denote visual tokens at a given feature level involved in post-fusion (in practice, we apply the same UGF to each of the three decoder input levels with shared parameters), and let $\mathbf{T} \in \mathbb{R}^{B \times L \times C}$ be the text tokens with padding mask $\mathbf{m}\in\{0,1\}^{B\times L}$. UGF computes cross-attention with visual tokens as queries and text tokens as keys/values, and obtains a visual increment via a projection:
\begin{equation}
	\Delta \mathbf{V} = \mathrm{Proj}\big(\mathrm{CrossAttn}(\mathbf{V}, \mathbf{T}, \mathbf{m})\big),
\end{equation}
where $\mathrm{CrossAttn}(\cdot)$ is multi-head attention and $\mathrm{Proj}(\cdot)$ corresponds to Linear+Dropout in implementation to stabilize the update scale.

Since RUS predicts uncertainty logits $U$ at stride-$32$ resolution, we align them to the token resolution of the fused level by bilinear resizing on logits and flatten:
\begin{equation}
	\begin{aligned}
		U^{*} &= \mathrm{Resize}\big(U, H_{*}, W_{*}\big), \\
		\mathbf{u} &= \mathrm{Flatten}\big(U^{*}\big) \in \mathbb{R}^{B \times N_{*} \times 1},
	\end{aligned}
\end{equation}
where $N_{*}=H_{*}W_{*}$. We map uncertainty to fusion coefficients using a Sigmoid with a learnable affine transform:
\begin{equation}
	\mathbf{g} = \sigma(\alpha \mathbf{u} + \beta),
\end{equation}
where $\alpha$ and $\beta$ are learnable scalars (initialized to encourage a monotonic mapping so that higher $U$ yields stronger fusion). Finally, UGF injects the gated update in a residual manner and applies layer normalization:
\begin{equation}
	\mathbf{V}^{+} = \mathrm{LN}(\mathbf{V} + \mathbf{g} \odot \Delta \mathbf{V}).
\end{equation}

This design explicitly converts uncertainty cues into fusion-strength control: positions with high uncertainty receive stronger textual constraints for disambiguation, whereas positions with low uncertainty suppress cross-modal perturbations to prevent the spread of linguistic noise. As an independent post-fusion module, UGF improves robustness under complex descriptions and highly confusable scenes without altering the backbone or the decoder interfaces.

\subsection{Uncertainty-Driven Local Refinement}
\label{sec:udlr}

Errors in remote-sensing oriented segmentation are typically concentrated around boundaries and fine-grained structures: targets are often slender, fragmented, or weak-contrast, and lie adjacent to background objects (e.g., along roads, shorelines, and building edges) \cite{refinedunet2020, yuan2023rrsis}. As a result, coarse predictions from the main decoder are prone to local adhesion, breakage, and boundary misalignment. To address these high-risk, shape-irregular local errors, we propose Uncertainty-Driven Local Refinement (UDLR): the uncertainty map determines \emph{where} secondary corrections are needed, and a continuous soft region replaces Top-$K$ square ROIs (as used in PointRend \cite{kirillov2020pointrend}), thereby reducing boundary artifacts and better matching the true geometry of remote-sensing targets.

Let the high-resolution mask feature be $\mathbf{F}$ (the highest-resolution decoder feature), the coarse foreground logit be $P_{\mathrm{fg}}$, and the uncertainty probability map be $U^{p}=\sigma(\tilde{U}) \in [0, 1]$ (obtained by applying Sigmoid to the aligned uncertainty logit $\tilde{U}$). UDLR first maps $U^{p}$ to a continuous soft refinement mask $\mathbf{M}$, which jointly represents the refinement region and its strength. We construct $\mathbf{M}$ via a temperature-scaled thresholded Sigmoid:
\begin{equation}
	\mathbf{M} = \sigma \left( \frac{\mathrm{Blur}(U^{p}) - \tau}{t} \right),
\end{equation}
where $\tau$ controls the central threshold of the refinement region and $t$ controls the smoothness of the transition; $\mathrm{Blur}(\cdot)$ is an optional local smoothing operator. To improve stability of region selection, $\mathbf{M}$ is used only as a selection/weighting signal and gradients are stopped through $\mathbf{M}$:
\begin{equation}
	\mathbf{M}\leftarrow \mathrm{StopGrad}(\mathbf{M}).
\end{equation}

Next, a lightweight convolutional refinement head $\mathcal{R}(\cdot)$ predicts a logit residual $\Delta$ at full resolution from the high-resolution feature, the coarse prediction, and uncertainty:
\begin{equation}
	\Delta = \mathcal{R} \left( [ \mathbf{F} ; P_{\mathrm{fg}} ; U^{p} ] \right),
\end{equation}
where $\mathcal{R}(\cdot)$ consists of two $3\times3$ convolution layers with ReLU activations, followed by a $1\times1$ convolution to output $\Delta$. To start from near-identity refinement and improve stability, the final $1\times1$ projection is initialized to produce near-zero residual. Finally, UDLR spatially weights the residual with the soft mask, correcting the coarse prediction only in high-uncertainty regions:
\begin{equation}
	P_{\mathrm{fg}}^{\mathrm{ref}} = P_{\mathrm{fg}} + \mathbf{M} \odot \Delta.
\end{equation}

Optionally, we apply auxiliary supervision on the refined prediction:
\begin{equation}
	\mathcal{L}_{\mathrm{ref}}=\mathcal{L}_{\mathrm{seg}}\big(\sigma(P_{\mathrm{fg}}^{\mathrm{ref}}), y\big),
\end{equation}
and the overall objective is
\begin{equation}
	\mathcal{L}=\mathcal{L}_{\mathrm{seg}}+\lambda_{\mathrm{unc}}\mathcal{L}_{\mathrm{unc}}+\lambda_{\mathrm{ref}}\mathcal{L}_{\mathrm{ref}},
\end{equation}
where $\lambda_{\mathrm{ref}}$ is set to $0$ when the auxiliary refinement supervision is disabled.

This design offers two key advantages: updates are explicitly confined to high-response regions of $\mathbf{M}$, reducing unnecessary perturbations in low-uncertainty areas; and refinement regions are adaptively generated from the spatial structure of $U^{p}$, naturally conforming to slender or irregular boundaries and avoiding hard-edge artifacts of fixed ROIs. By sharing the same prior $\tilde{U}$ (logit) / $U^{p}$ (probability) with uncertainty modeling, UDLR aligns refinement with the model's risk assessment, thereby consistently improving segmentation quality around boundaries and fine details.

\subsection{Computation Overhead}
\label{sec:overhead}

The proposed modules introduce lightweight overhead on top of the two-tower backbone.
RUS consists of one cross-modal alignment step and a shallow MLP for token-wise uncertainty prediction at the coarsest feature level.
UGF adds a single post-fusion cross-attention update with a simple scalar-gated modulation (applied once before decoding), and therefore minimally increases computation.
UDLR employs a shallow convolutional residual predictor and applies corrections through an uncertainty-derived soft mask, focusing effective updates on uncertain regions.
Overall, the method remains practical for remote-sensing referring segmentation while improving robustness under ambiguous referring conditions and fine-structure errors.

\section{Experiments}
\label{section:EXPERIMENTS}

\subsection{Datasets and Evaluation Metrics}

\subsubsection{Datasets}
To comprehensively evaluate the effectiveness and robustness of the proposed framework, we conduct experiments on three challenging remote sensing referring segmentation benchmarks: \textbf{RefSegRS} \cite{yuan2023rrsis}, \textbf{RRSIS-D} \cite{han2023rrsisd}, and \textbf{RISBench} \cite{zhan2024rsvg}. These datasets are selected to cover a wide range of difficulties, including high-resolution targets, complex semantic attributes, and large-scale resolution variations. The detailed statistics and data partition for each dataset are summarized in Table \ref{tab:datasets}.

\begin{table}[h]
	\centering
	\caption{Statistical Details of the Three Datasets Used in Our Experiments.}
	\label{tab:datasets}
	\setlength{\tabcolsep}{5pt}
	\renewcommand{\arraystretch}{1.2}
	\begin{tabular}{l|c|c|ccc}
		\hline
		\multirow{2}{*}{\textbf{Dataset}} & \multirow{2}{*}{\textbf{Resolution}} & \multirow{2}{*}{\textbf{Image Size}} & \multicolumn{3}{c}{\textbf{Split (Image-Text Pairs)}} \\ \cline{4-6} 
		&  &  & \textbf{Train} & \textbf{Val} & \textbf{Test} \\ \hline
		RefSegRS & $0.13$~m & $512 \times 512$ & 2,172 & 431 & 1,817 \\
		RRSIS-D & - & $800 \times 800$ & 12,181 & 1,740 & 3,481 \\
		RISBench & $0.1$~m - $30$~m & $512 \times 512$ & 26,300 & 10,013 & 16,159 \\ \hline
	\end{tabular}
\end{table}

\begin{itemize}
	\item \textbf{RefSegRS \cite{yuan2023rrsis}:} As a high-resolution benchmark, RefSegRS is characterized by its fine-grained imagery (0.13m spatial resolution). It poses significant challenges for precise boundary delineation of objects in complex scenes.
	
	\item \textbf{RRSIS-D \cite{han2023rrsisd}:} This dataset distinguishes itself with rich semantic annotations. Unlike standard datasets, it includes 20 object categories and 7 specific attributes, requiring the model to capture intricate linguistic cues and geometric details (e.g., orientation).
	
	\item \textbf{RISBench \cite{zhan2024rsvg}:} Being a large-scale dataset, RISBench introduces extreme scale variations with spatial resolutions spanning from $0.1$~m to $30$~m. This diversity tests the model's capability to handle objects with significant size differences across varying observation altitudes.
\end{itemize}

\subsubsection{Evaluation Metrics}
Following the standard evaluation protocols in referring segmentation \cite{yang2022lavt, yuan2023rrsis}, we adopt Overall Intersection-over-Union (oIoU), Mean Intersection-over-Union (mIoU), and Precision at thresholds $\text{Pr}@X$ ($X \in \{0.5, 0.6, 0.7, 0.8, 0.9\}$) as our primary metrics.

\textbf{oIoU} focuses on the overall pixel-level accuracy and is calculated by accumulating the intersection and union areas over the entire dataset. It gives more weight to large-scale objects:
\begin{equation}
	\text{oIoU} = \frac{\sum_{i=1}^{N} \mathcal{I}_i}{\sum_{i=1}^{N} \mathcal{U}_i},
\end{equation}
where $N$ is the total number of test samples, and $\mathcal{I}_i$ and $\mathcal{U}_i$ represent the intersection and union areas between the predicted mask and the ground truth for the $i$-th sample, respectively.

\textbf{mIoU} provides a macro-average performance by treating each sample equally, regardless of the object size:
\begin{equation}
	\text{mIoU} = \frac{1}{N} \sum_{i=1}^{N} \frac{\mathcal{I}_i}{\mathcal{U}_i}.
\end{equation}

Additionally, \textbf{Pr@X} measures the percentage of test samples where the IoU score exceeds the threshold $X$, reflecting the reliability of the model under different strictness levels.

\subsection{Experimental Setup}

\subsubsection{Network Configuration}
Our framework is implemented on the PyTorch platform \cite{paszke2019pytorch}. To ensure comprehensive feature extraction, we employ ConvNeXt (ConvNeXt-Base) \cite{liu2022convnet}. Specifically, ConvNeXt-Base utilizes weights derived from the self-supervised SMLFR algorithm (Generative ConvNet Foundation Model) \cite{dong2024generative}, which is designed to capture sparse and low-frequency structures in remote sensing images. For linguistic processing, we adopt the standard 12-layer BERT-base model \cite{devlin2018bert} with a hidden dimension of 768. The input images are uniformly resized to $480 \times 480$ pixels, and the maximum length for textual descriptions is truncated to 20 tokens.

\subsubsection{Training Protocol}
The model is trained on a cluster of eight NVIDIA A800 GPUs with a total batch size of 32. We utilize the AdamW optimizer \cite{loshchilov2017decoupled} with an initial learning rate of $5 \times 10^{-5}$ and a weight decay coefficient of 0.01. The learning rate follows a polynomial decay schedule. The entire training process spans 40 epochs. Consistent with prior works, no additional data augmentation or post-processing techniques are applied during the training and inference phases.

\subsection{Comparison with State-of-the-art Methods}

\subsubsection{RRSIS-D}

To evaluate the effectiveness of our proposed method, we conducted extensive experiments on the RRSIS-D dataset, and the comparison results are reported in Table~\ref{rrsisd_comparison}. RRSIS-D provides rich semantic annotations with \textbf{20 object categories} and \textbf{7 specific attributes}, where referring expressions frequently involve fine-grained linguistic cues and geometric details (e.g., orientation and spatial relations) \cite{han2023rrsisd}. Such characteristics require the model to align language with visual evidence more precisely, rather than relying solely on coarse category recognition.

We compared our framework with a wide range of methods, including classical LSTM-based models (e.g., RRN~\cite{li2018referring}, CMPC~\cite{huang2020referring}) and recent CLIP/BERT-based approaches (e.g., CRIS~\cite{wang2022cris}, LAVT~\cite{yang2022lavt}, RIS-DMMI~\cite{hu2023beyond}, robust-ref-seg~\cite{wu2024towards}, and CARIS~\cite{liu2023caris}). Overall, LSTM-based methods generally underperform compared to CLIP- and BERT-based solutions. This is mainly because RRSIS-D descriptions often contain attribute-level and relation-level constraints, for which capturing global semantic dependencies is crucial. Consequently, modern combinations of hierarchical visual encoders (e.g., Swin Transformer \cite{liu2021swin} or ConvNeXt \cite{liu2022convnet}) and BERT-style text encoders constitute strong baselines on this dataset.

Among all compared methods, our framework achieves superior performance. The baseline CroBIM (with ConvNeXt-B) already delivers competitive results and surpasses the previous state-of-the-art method CARIS \cite{liu2023caris} in terms of mIoU. More importantly, CroBIM-U further improves performance by explicitly addressing \textbf{referential uncertainty}, which is particularly relevant to RRSIS-D where multiple instances may share similar categories but differ in attributes (e.g., ``the rotated/leftmost/elongated object''). By reducing ambiguity in language-to-region matching, CroBIM-U achieves the best results across almost all metrics.

Specifically, CroBIM-U achieves a \textbf{Pr@0.5} of \textbf{76.31\%} on the validation set and \textbf{75.60\%} on the test set, outperforming CARIS (71.61\%/71.50\%) by over \textbf{4\%}. In terms of \textbf{mIoU}, which better reflects segmentation quality across diverse categories and attribute-conditioned targets, CroBIM-U reaches \textbf{66.07\%} (Val) and \textbf{65.07\%} (Test), bringing substantial gains of \textbf{+3.19\%} and \textbf{+2.95\%} over CARIS, respectively.

Although robust-ref-seg \cite{wu2024towards} reports a slightly higher oIoU on the test set, CroBIM-U achieves the highest oIoU on the validation split (\textbf{77.83\%}). The combination of \textbf{higher mIoU} and \textbf{competitive oIoU} suggests that our method is particularly effective in improving the segmentation of attribute-sensitive and potentially smaller or visually similar targets, rather than being dominated by large-area classes. These results validate the effectiveness of our design in achieving precise pixel-level localization under rich semantic and geometric constraints on RRSIS-D.

\begin{table*}[tbp]
	\centering
	\caption{Comparison with state-of-the-art methods on the RRSIS-D dataset. Optimal and sub-optimal performance in each metric are marked by \textcolor{red}{\textbf{red}} and \textcolor{blue}{\textbf{blue}}.}
	\label{rrsisd_comparison}
	\renewcommand\arraystretch{1.0} 
	\setlength{\tabcolsep}{3.5pt}
	\fontsize{8}{11}\selectfont
	
	\resizebox{\textwidth}{!}{
		\begin{tabular}{l|c|c|cc|cc|cc|cc|cc|cc|cc}
			\hline
			\multirow{2}{*}{Method} & \multirow{2}{*}{Visual Encoder} & \multirow{2}{*}{Text Encoder} & \multicolumn{2}{c|}{Pr@0.5} & \multicolumn{2}{c|}{Pr@0.6} & \multicolumn{2}{c|}{Pr@0.7} & \multicolumn{2}{c|}{Pr@0.8} & \multicolumn{2}{c|}{Pr@0.9} & \multicolumn{2}{c|}{oIoU} & \multicolumn{2}{c}{mIoU} \\ \cline{4-17} 
			& & & Val & Test & Val & Test & Val & Test & Val & Test & Val & Test & Val & Test & Val & Test \\ \hline
			RRN\cite{li2018referring} & ResNet-101 & LSTM & 51.09 & 51.07 & 42.47 & 42.11 & 33.04 & 32.77 & 20.80 & 21.57 & 6.14 & 6.37 & 66.53 & 66.43 & 46.06 & 45.64 \\
			CSMA\cite{ye2019cross} & ResNet-101 & None & 55.68 & 55.32 & 48.04 & 46.45 & 38.27 & 37.43 & 26.55 & 25.39 & 9.02 & 8.15 & 69.68 & 69.43 & 48.85 & 48.54 \\
			LSCM\cite{hui2020linguistic} & ResNet-101 & LSTM & 57.12 & 56.02 & 48.04 & 46.25 & 37.87 & 37.70 & 26.35 & 25.28 & 7.93 & 7.86 & 69.28 & 69.10 & 50.36 & 49.92 \\
			CMPC\cite{huang2020referring} & ResNet-101 & LSTM & 57.93 & 55.83 & 48.85 & 47.40 & 36.94 & 35.28 & 25.25 & 25.45 & 9.31 & 9.20 & 70.15 & 69.41 & 51.01 & 49.24 \\
			BRINet\cite{hu2020bi} & ResNet-101 & LSTM & 58.79 & 56.90 & 49.54 & 48.77 & 39.65 & 38.61 & 28.21 & 27.03 & 9.19 & 8.93 & 70.73 & 69.68 & 51.41 & 49.45 \\
			CMPC+\cite{liu2021cross} & ResNet-101 & LSTM & 59.19 & 57.95 & 49.41 & 48.31 & 38.67 & 37.61 & 25.91 & 24.33 & 8.16 & 7.94 & 70.80 & 70.13 & 51.63 & 50.12 \\    
			BKINet\cite{ding2023bilateral} & ResNet-101 & CLIP & 58.79 & 56.90 & 49.54 & 48.77 & 39.65 & 39.12 & 28.21 & 27.03 & 9.19 & 9.16 & 70.78 & 69.89 & 51.14 & 49.65 \\
			ETRIS\cite{xu2023bridging} & ResNet-101 & CLIP & 62.10 & 61.07 & 53.73 & 50.99 & 43.12 & 40.94 & 30.79 & 29.30 & 12.90 & 11.43 & 72.75 & 71.06 & 55.21 & 54.21 \\
			CRIS\cite{wang2022cris} & ResNet-101 & CLIP & 56.44 & 54.84 & 47.87 & 46.77 & 39.77 & 38.06 & 29.31 & 28.15 & 11.84 & 11.52 & 70.98 & 70.46 & 50.75 & 49.69 \\
			LGCE\cite{yuan2024rrsis} & Swin-B & BERT & 68.10 & 67.65 & 60.61 & 61.53 & 51.45 & 51.42 & 42.34 & 39.62 & 23.85 & 22.94 & 76.68 & 76.33 & 60.16 & 59.37 \\
			LAVT\cite{yang2022lavt} & Swin-B & BERT & 65.23 & 63.98 & 58.79 & 57.57 & 50.29 & 49.30 & 40.11 & 38.06 & 23.05 & 22.29 & 76.27 & 76.16 & 57.72 & 56.82 \\
			RMSIN\cite{liu2024rotated} & Swin-B & BERT & 68.39 & 67.16 & 61.72 & 60.36 & 52.24 & 50.16 & 41.44 & 38.72 & 23.16 & 22.81 & \textcolor{blue}{\textbf{77.53}} & 75.79 & 60.23& 58.79 \\
			CrossVLT\cite{cho2023cross} & Swin-B & BERT & 67.07 & 66.42 & 59.54 & 59.41 & 50.80 & 49.76 & 40.57 & 38.67 & 23.51 & 23.30 & 76.25 & 75.48 & 59.78 & 58.48 \\
			RIS-DMMI\cite{hu2023beyond} & Swin-B & BERT & 70.40 & 68.74 & 63.05 & 60.96 & 54.14 & 50.33 & 41.95 & 38.38 & 23.85 & 21.63 & 77.01 & 76.20 & 61.70 & 60.25 \\
			robust-ref-seg\cite{wu2024towards} & Swin-B & BERT & 64.22 & 66.59 & 58.72 & 59.58 & 50.00 & 49.93 & 35.78 & 38.72 & 24.31 & 23.30 & 76.39 & \textcolor{red}{\textbf{77.40}} & 58.92 & 58.91 \\
			CARIS\cite{liu2023caris} & Swin-B & BERT & 71.61 & 71.50 & 64.66 & 63.52 & 54.14 & 52.92 & 42.76 & 40.94 & 23.79 & \textcolor{blue}{\textbf{23.90}} & 77.48 & \textcolor{blue}{\textbf{77.17}} & 62.88 & 62.12 \\
			CroBIM & Swin-B & BERT & 74.20 & \textcolor{blue}{\textbf{75.00}} & 66.15 & 66.32 & 54.08 & 54.31 & 41.38 & 41.09 & 22.30 & 21.78 & 76.24 & 76.37 & 63.99 & 64.24\\  
			CroBIM & ConvNeXt-B & BERT & \textcolor{blue}{\textbf{74.94}} & 74.58 & \textcolor{blue}{\textbf{67.64}} & \textcolor{blue}{\textbf{67.57}} & \textcolor{blue}{\textbf{57.18}} & \textcolor{blue}{\textbf{55.59}} & \textcolor{blue}{\textbf{44.66}} & \textcolor{blue}{\textbf{41.63}} & \textcolor{blue}{\textbf{24.60}} & 23.56 & 76.94 & 75.99 & \textcolor{blue}{\textbf{65.05}} & \textcolor{blue}{\textbf{64.46}} \\ \hline
			CroBIM-U & ConvNeXt-B & BERT & \textcolor{red}{\textbf{76.31}} & \textcolor{red}{\textbf{75.60}} & \textcolor{red}{\textbf{68.30}} & \textcolor{red}{\textbf{67.68}} & \textcolor{red}{\textbf{57.67}} & \textcolor{red}{\textbf{56.47}} & \textcolor{red}{\textbf{44.97}} & \textcolor{red}{\textbf{42.57}} & \textcolor{red}{\textbf{24.89}} & \textcolor{red}{\textbf{24.16}} & \textcolor{red}{\textbf{77.83}} & 76.70 & \textcolor{red}{\textbf{66.07}} & \textcolor{red}{\textbf{65.07}} \\ \hline
		\end{tabular}
	}
	
	\vspace{1.5em}

	\makeatletter
	\def\@captype{figure}
	\makeatother
	\includegraphics[width=1\linewidth]{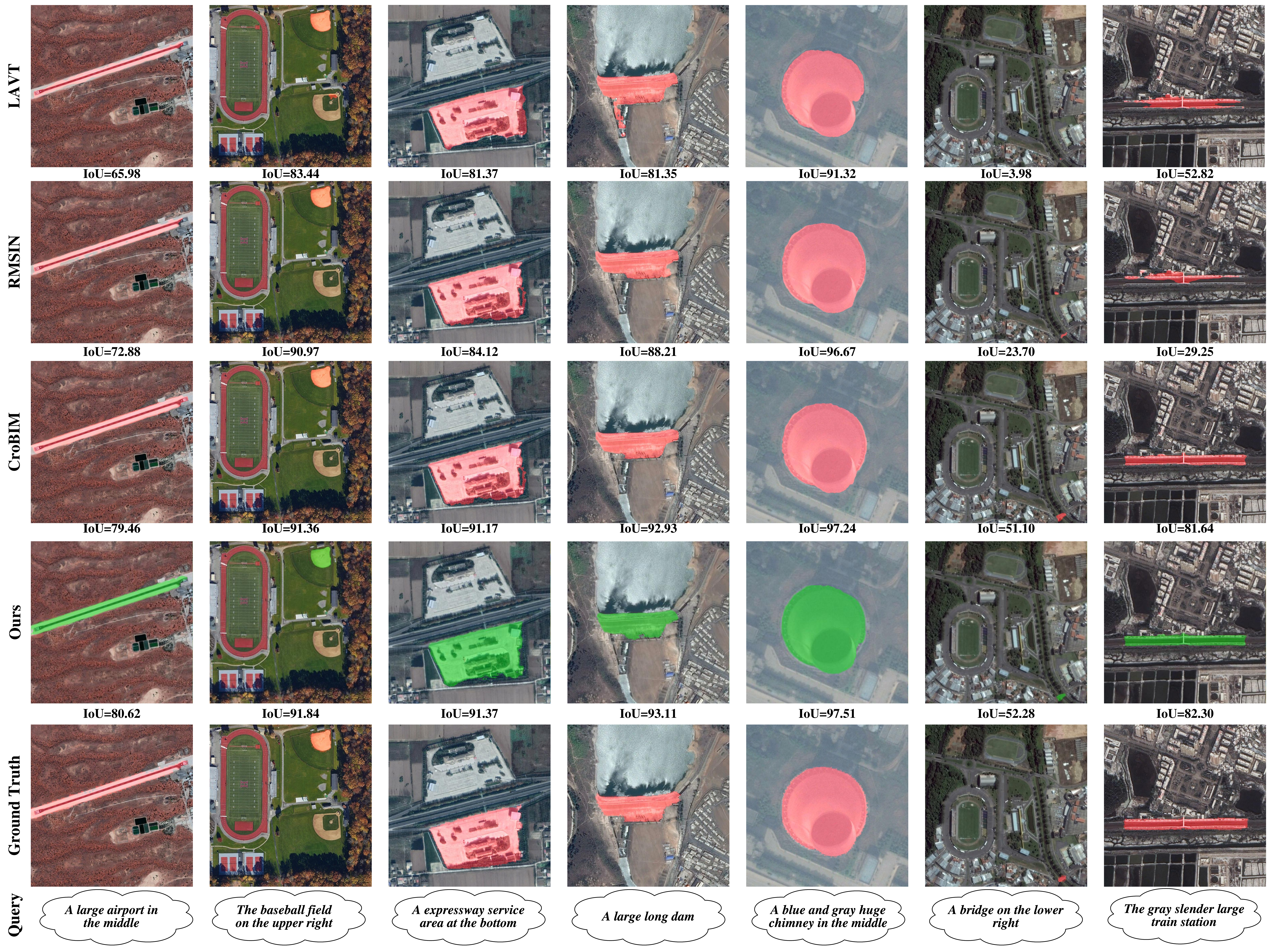}
	\caption{Visualization of segmentation results for CroBIM and comparison methods on the RRSIS-D dataset test set, with corresponding IoU scores displayed.}
	\label{vis_rrsisd}
\end{table*}

\subsubsection{RefSegRS}

We conduct experiments on the RefSegRS dataset, and the results on both the validation and test sets are reported in Table~\ref{refsegrs_comparison}. RefSegRS is a high-resolution benchmark with fine-grained imagery (0.13m spatial resolution), where objects often exhibit subtle boundaries and appear in cluttered scenes \cite{yuan2023rrsis}. Therefore, metrics under stricter overlap thresholds (e.g., Pr@0.7--Pr@0.9) and IoU-based scores are particularly indicative of a model's capability in accurate boundary delineation.

As shown in Table.~\ref{refsegrs_comparison}, \textbf{CroBIM-U} achieves the best performance across all precision thresholds. On the test set, it attains \textbf{76.68\%} (Pr@0.5) and \textbf{62.54\%} (Pr@0.6), surpassing previous strong competitors such as RIS-DMMI \cite{hu2023beyond} (63.89\% / 44.30\%). More importantly, CroBIM-U maintains a clear advantage under more stringent criteria that emphasize boundary quality: it reaches \textbf{34.81\%} (Pr@0.7), \textbf{13.50\%} (Pr@0.8), and \textbf{3.26\%} (Pr@0.9) on the test split. These gains under high overlap thresholds demonstrate that CroBIM-U produces masks with tighter alignment to fine object contours, which is crucial for this high-resolution dataset.

CroBIM-U also delivers the best IoU-based results, achieving \textbf{73.81\%} oIoU and \textbf{60.08\%} mIoU on the test set, indicating improved overall mask completeness and robustness in complex scenes. Compared with CroBIM using the same ConvNeXt-B backbone, CroBIM-U further improves the test performance from 75.89\% to 76.68\% on Pr@0.5 and from 61.42\% to 62.54\% on Pr@0.6, while boosting oIoU by \textbf{+1.48\%}. These consistent improvements, especially under strict overlap thresholds, confirm the superiority of CroBIM-U in fine-grained boundary delineation on RefSegRS.

\begin{table*}[htbp]
	\centering

	\caption{Comparison with state-of-the-art methods on the RefSegRS dataset. Optimal and sub-optimal performance in each metric are marked by \textcolor{red}{\textbf{red}} and \textcolor{blue}{\textbf{blue}}.}
	\label{refsegrs_comparison}
	\renewcommand\arraystretch{1.0} 
	\setlength{\tabcolsep}{3pt}   
	\fontsize{8}{11}\selectfont
	
	\resizebox{\textwidth}{!}{
		\begin{tabular}{l|c|c|cc|cc|cc|cc|cc|cc|cc}
			\hline
			\multirow{2}{*}{Method} & \multirow{2}{*}{Visual Encoder} & \multirow{2}{*}{Text Encoder} & \multicolumn{2}{c|}{Pr@0.5} & \multicolumn{2}{c|}{Pr@0.6} & \multicolumn{2}{c|}{Pr@0.7} & \multicolumn{2}{c|}{Pr@0.8} & \multicolumn{2}{c|}{Pr@0.9} & \multicolumn{2}{c|}{oIoU} & \multicolumn{2}{c}{mIoU} \\ \cline{4-17} 
			& & & Val & Test & Val & Test & Val & Test & Val & Test & Val & Test & Val & Test & Val & Test \\ \hline
			RRN\cite{li2018referring} & ResNet-101 & LSTM & 55.43 & 30.26 & 42.98 & 23.01 & 23.11 & 14.87 & 13.72 & 7.17 & 2.64 & 0.98 & 69.24 & 65.06 &  50.81 & 41.88 \\
			CMSA\cite{ye2019cross} & ResNet-101 & None & 39.24 & 26.14 & 38.44 & 18.52 & 20.39  & 10.66 & 11.79 & 4.71 & 1.52 & 0.69 & 63.84 & 62.11 & 43.62 & 38.72 \\
			LSCM\cite{hui2020linguistic} & ResNet-101 & LSTM  & 56.82 & 31.54 & 41.24 & 20.41 & 21.85 & 9.51 & 12.11 & 5.29 & 2.51 & 0.84 & 62.82 & 61.27 & 40.59 & 35.54 \\
			BRINet\cite{hu2020bi} & ResNet-101 & LSTM & 36.86 & 20.72 & 35.53 & 14.26 & 19.93 & 9.87 & 10.66 & 2.98 & 2.84 & 1.14 & 61.59 & 58.22 & 38.73  & 31.51 \\
			MAttNet\cite{yu2018mattnet} & ResNet-101 & LSTM & 48.56 & 28.79 & 40.26 & 22.51 & 20.59 & 11.32 & 12.98 & 3.62 & 2.02 & 0.79 & 66.84 & 64.28 & 41.73 & 33.42 \\
			BKINet\cite{ding2023bilateral} & ResNet-101 & CLIP & 52.04 & 36.12 & 35.31 & 20.62 & 18.35 & 15.22 & 12.78 & 6.26 & 1.23 & 1.33 & 75.37 & 63.37 & 56.12 & 40.41 \\
			ETRIS\cite{xu2023bridging} & ResNet-101 & CLIP & 54.99 & 35.77 & 35.03 & 23.00 & 25.06 & 13.98 & 12.53 & 6.44 & 1.62 & 1.10 & 72.89 & 65.96 & 54.03 & 43.11 \\
			CRIS\cite{wang2022cris} & ResNet-101 & CLIP & 53.13 & 35.77 & 36.19 & 24.11 & 24.36 & 14.36 & 11.83 & 6.38 & 2.55 & 1.21 & 72.14 & 65.87 & 53.74 & 43.26 \\
			RMSIN\cite{liu2024rotated} & Swin-B & BERT & 68.21 & 42.32 & 46.64 & 25.87 & 24.13 & 14.20 & 13.69 & 6.77 & 3.25 & 1.27 & 74.40 & 68.31 & 54.24 & 42.63 \\
			CrossVLT\cite{cho2023cross} & Swin-B & BERT & 67.52 & 41.94 & 43.85 & 25.43 & 25.99 & 15.19 & 14.62 & 3.71 & 1.87 & 1.76 & 76.12 & 69.73 & 55.27 & 42.81 \\
			RIS-DMMI\cite{hu2023beyond} & Swin-B & BERT & 86.17 & 63.89 & 74.71 & 44.30 & 38.05 & 19.81 & 18.10 & 6.49 & 3.25 & 1.00 & 74.02 & 68.58 & 65.72 & 52.15 \\
			CARIS\cite{liu2023caris} & Swin-B & BERT & 68.45 & 45.40 & 47.10 & 27.19 & 25.52 & 15.08 & 14.62 & 7.87 & 3.71 & 1.98 & 75.79 & 69.74 & 54.30 & 42.66 \\
			robust-ref-seg\cite{wu2024towards} & Swin-B & BERT & 81.67 & 50.25 & 52.44 & 28.01 & 30.86 & 17.83 & 17.17 & 9.19 & 5.80 & 2.48 & 77.74 & 71.13 & 60.44 & 47.12 \\
			LGCE\cite{yuan2024rrsis} & Swin-B & BERT & 79.81 & 50.19 & 54.29 & 28.62 & 29.70 & 17.17 & 15.31 & 9.36 & 5.10 & 2.15 & 78.24 & 71.59 & 60.66 & 46.57 \\ 
			LAVT\cite{yang2022lavt} & Swin-B & BERT & 80.97 & 51.84 & 58.70 & 30.27 & 31.09 & 17.34 & 15.55 & 9.52 & 4.64 & 2.09 & \textcolor{blue}{\textbf{78.50}} & 71.86 & 61.53 & 47.40 \\ 
			CroBIM & Swin-B & BERT & 87.24 & 64.83 & 75.17 & 44.41 & 44.78 & 17.28 & 19.03 & 9.69 & \textcolor{blue}{\textbf{6.26}} & 2.20 &  \textcolor{red}{\textbf{78.85}} & 72.30 & 65.79 & 52.69 \\     
			CroBIM & ConvNeXt-B & BERT & \textcolor{blue}{\textbf{93.04}} & \textcolor{blue}{\textbf{75.89}} & \textcolor{blue}{\textbf{87.70}} & \textcolor{blue}{\textbf{61.42}} & \textcolor{blue}{\textbf{66.13}} & \textcolor{blue}{\textbf{34.07}} & \textcolor{blue}{\textbf{26.91}} & \textcolor{blue}{\textbf{12.99}} & 5.80 & \textcolor{blue}{\textbf{2.75}} & 77.95 & \textcolor{blue}{\textbf{72.33}} & \textcolor{blue}{\textbf{71.93}} & \textcolor{blue}{\textbf{59.77}} \\     
			\hline
			CroBIM-U & ConvNeXt-B & BERT & \textcolor{red}{\textbf{94.11}} & \textcolor{red}{\textbf{76.68}} & \textcolor{red}{\textbf{88.54}} & \textcolor{red}{\textbf{62.54}} & \textcolor{red}{\textbf{68.45}} & \textcolor{red}{\textbf{34.81}} & \textcolor{red}{\textbf{27.22}} & \textcolor{red}{\textbf{13.50}} & \textcolor{red}{\textbf{6.57}} & \textcolor{red}{\textbf{3.26}} & 78.03 & \textcolor{red}{\textbf{73.81}} & \textcolor{red}{\textbf{72.55}} & \textcolor{red}{\textbf{60.08}} \\     
			\hline
		\end{tabular}
	}
	
	\vspace{1.5em}

	\makeatletter
	\def\@captype{figure}
	\makeatother
	\centering
	\includegraphics[width=1\linewidth]{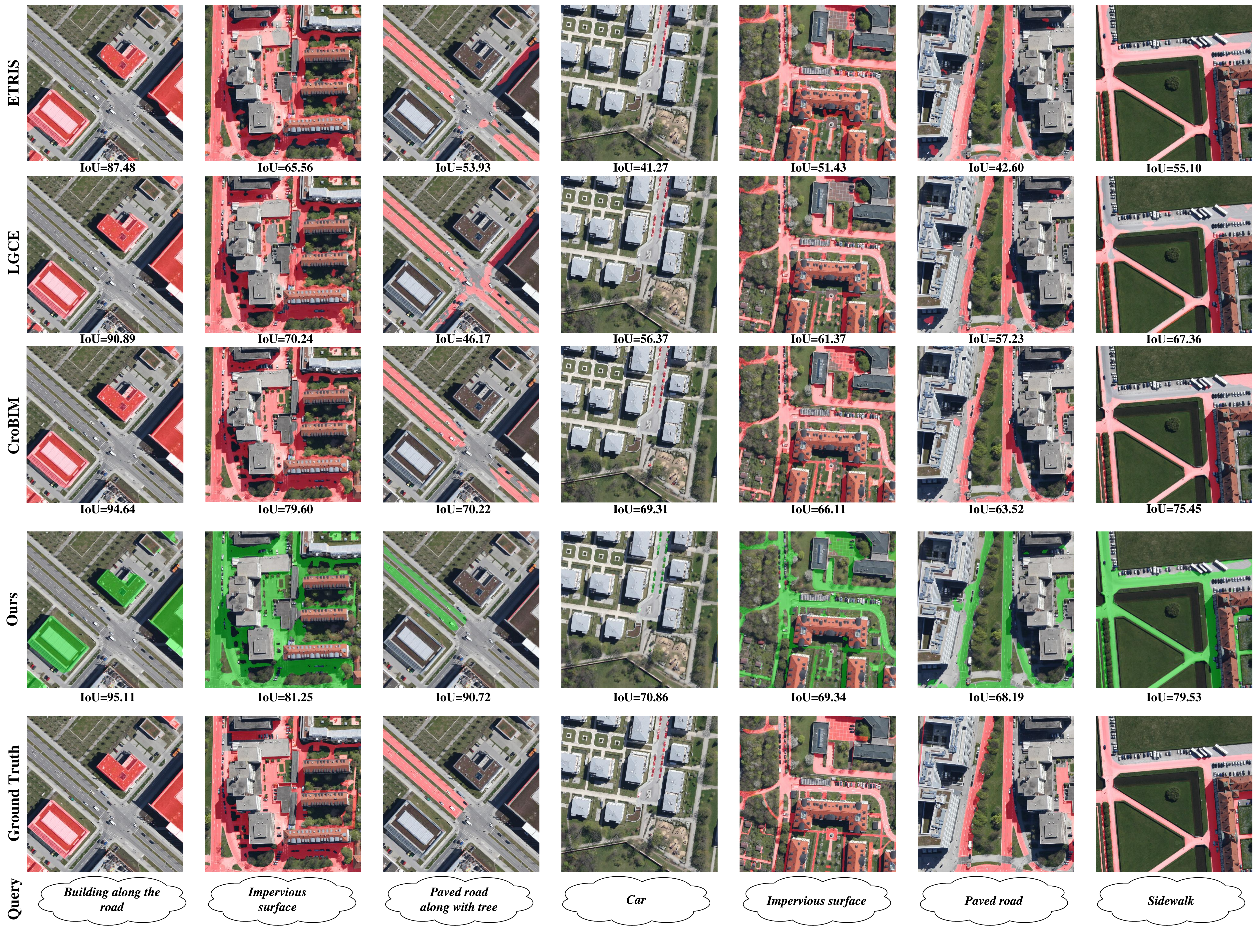}
	\caption{Visualization of segmentation results for CroBIM and comparison methods on the RefSegRS dataset test set, with corresponding IoU scores displayed.}
	\label{vis_refsegrs}
\end{table*}

\subsubsection{RISBench}

We conduct experiments on the RISBench dataset and report the results on both the validation and test sets in Table ~\ref{risbench_comparison}. As a large-scale benchmark with \textbf{extreme scale variations} (spatial resolutions ranging from \textbf{0.1m to 30m}) \cite{zhan2024rsvg}, RISBench poses substantial challenges to referring segmentation models: targets may appear with drastically different sizes and visual patterns across observation altitudes, making robust multi-scale localization and precise mask prediction particularly critical.

As shown in Table.~\ref{risbench_comparison}, \textbf{CroBIM-U} achieves the \textbf{best performance across all precision thresholds (Pr@0.5--Pr@0.9)} and delivers the highest \textbf{mIoU} on both splits. On the test set, CroBIM-U reaches \textbf{77.73\%} (Pr@0.5) and \textbf{73.40\%} (Pr@0.6), exceeding strong BERT-based competitors such as CARIS \cite{liu2023caris} (73.94\% / 68.93\%) and CrossVLT \cite{cho2023cross} (70.62\% / 65.05\%). More importantly, under stricter overlap constraints that better reflect fine localization quality across varying object scales, CroBIM-U maintains clear advantages, achieving \textbf{67.05\%} (Pr@0.7), \textbf{56.85\%} (Pr@0.8), and \textbf{35.52\%} (Pr@0.9) on the test set. The improvements at high thresholds indicate that CroBIM-U is more capable of producing tightly aligned masks even when targets are small, thin, or visually ambiguous due to resolution changes.

In terms of overall segmentation quality, CroBIM-U obtains the highest \textbf{mIoU} of \textbf{69.15\%} (Val) and \textbf{69.62\%} (Test), demonstrating stronger category-balanced performance on this large-scale dataset with substantial size diversity. Compared with CroBIM using the same ConvNeXt-B backbone, CroBIM-U consistently improves high-threshold precision (e.g., \textbf{+1.45\%} on Pr@0.9 for the test split) and further boosts mIoU (\textbf{+0.45\%} Val / \textbf{+0.29\%} Test), validating that resolving \textbf{referential uncertainty} is beneficial when targets exhibit large scale fluctuations and similar instances may coexist across different resolutions. 

Notably, CroBIM-U shows \textbf{comparable oIoU} to CroBIM on RISBench, while significantly improving mIoU and high-threshold precision. This is expected because oIoU is more influenced by large-area regions, whereas RISBench contains many small-to-medium targets whose accurate segmentation is better reflected by \textbf{mIoU} and stricter \textbf{Pr} metrics. Overall, these results confirm that CroBIM-U offers stronger robustness to scale variation and yields more precise pixel-level localization on RISBench.

-
\begin{table*}[htbp]
	\centering

	\caption{Comparison with state-of-the-art methods on the RISBench dataset. Optimal and sub-optimal performance in each metric are marked by \textcolor{red}{\textbf{red}} and \textcolor{blue}{\textbf{blue}}.}
	\label{risbench_comparison}
	\renewcommand\arraystretch{1.0} 
	\setlength{\tabcolsep}{3pt}
	\fontsize{8}{11}\selectfont
	
	\resizebox{\textwidth}{!}{
		\begin{tabular}{l|c|c|cc|cc|cc|cc|cc|cc|cc}
			\hline
			\multirow{2}{*}{Method} & \multirow{2}{*}{Visual Encoder} & \multirow{2}{*}{Text Encoder} & \multicolumn{2}{c|}{Pr@0.5} & \multicolumn{2}{c|}{Pr@0.6} & \multicolumn{2}{c|}{Pr@0.7} & \multicolumn{2}{c|}{Pr@0.8} & \multicolumn{2}{c|}{Pr@0.9} & \multicolumn{2}{c|}{oIoU} & \multicolumn{2}{c}{mIoU} \\ \cline{4-17} 
			& & & Val & Test & Val & Test & Val & Test & Val & Test & Val & Test & Val & Test & Val & Test \\ \hline
			RRN\cite{li2018referring} & ResNet-101 & LSTM & 54.62 & 55.04 & 46.88 & 47.31 & 39.57 & 39.86 & 32.64 & 32.58 & 11.57 & 13.24 & 47.28 & 49.67 & 42.65 & 43.18 \\
			LSCM\cite{hui2020linguistic} & ResNet-101 & LSTM  & 55.87 & 55.26 & 47.24 & 47.14 & 40.22 & 40.10 & 33.55 & 33.29 & 12.78 & 13.91 & 47.99 & 50.08 & 43.21 & 43.69 \\
			BRINet\cite{hu2020bi} & ResNet-101 & LSTM & 52.11 & 52.87 & 45.17 & 45.39 & 37.98 & 38.64 & 30.88 & 30.79 & 10.28 & 11.86 & 46.27 & 48.73 & 41.54 & 42.91 \\
			MAttNet\cite{yu2018mattnet} & ResNet-101 & LSTM & 56.77 & 56.83 & 48.51 & 48.02 & 41.53 & 41.75 & 34.33 & 34.18 & 13.84 & 15.26 & 48.66 & 51.24 & 44.28 & 45.71 \\
			CMPC\cite{huang2020referring} & ResNet-101 & LSTM & 54.89 & 55.17 & 47.77 & 47.84 & 40.38 & 40.28 & 32.89 & 32.87 & 12.63 & 14.55 & 47.59 & 50.24 & 42.83 & 43.82\\
			CMPC+\cite{liu2021cross} & ResNet-101 & LSTM & 57.84 & 58.02 & 49.24 & 49.00 & 42.34 & 42.53 & 35.77 & 35.26 & 14.55 & 17.88 & 50.29 & 53.98 & 45.81 & 46.73 \\    
			LAVT\cite{yang2022lavt} & Swin-B & BERT & 68.27 & 69.40 & 62.71 & 63.66 & 54.46 & 56.10 & 43.13 & 44.95 & 21.61 & 25.21 & 69.39 & 74.15 & 60.45 & 61.93 \\
			RMSIN\cite{liu2024rotated} & Swin-B & BERT & 70.05 & 71.01 & 64.64 & 65.46 & 56.37 & 57.69 & 44.14 & 45.50 & 21.40 & 25.92 & 69.51 & 74.09 & 61.78 & 63.07 \\
			LGCE\cite{yuan2024rrsis} & Swin-B & BERT & 68.20 & 69.64 & 62.91 & 64.07 & 55.01 & 56.26 & 43.38 & 44.92 & 21.58 & 25.74 & 68.81 & 73.87 & 60.44 & 62.13 \\ 
			CrossVLT\cite{cho2023cross} & Swin-B & BERT & 70.05 & 70.62 & 64.29 & 65.05 & 56.97 & 57.40 & 44.49 & 45.80 & 21.47 & 26.10 & 69.77 & 74.33 & 61.54 & 62.84 \\ 
			CARIS\cite{liu2023caris} & Swin-B & BERT & 73.46 & 73.94 & 68.51 & 68.93 & 60.92 & 62.08 & 48.47 & 50.31 & 24.98 & 29.08 & \textcolor{blue}{\textbf{70.55}} & \textcolor{red}{\textbf{75.10}} & 64.40 & 65.79 \\
			RIS-DMMI\cite{hu2023beyond} & Swin-B & BERT & 71.27 & 72.05 & 66.02 & 66.48 & 58.22 & 59.07 & 45.57 & 47.16 & 22.43 & 26.57 & \textcolor{red}{\textbf{70.58}} & \textcolor{blue}{\textbf{74.82}} & 62.62 & 63.93 \\
			robust-ref-seg\cite{wu2024towards} & Swin-B & BERT & 67.42 & 69.15 & 61.72 & 63.24 & 53.64 & 55.33 & 40.71 & 43.27 & 19.43 & 24.20 & 69.50 & 74.23 & 59.37 & 61.25 \\
			CroBIM & Swin-B & BERT & 76.59 & 75.75 & 71.73 & 70.34 & 64.32 & 63.12 & 53.18 & 51.12 & 28.53 & 28.45 & 69.08 & 73.61 & 67.52 & 67.32 \\    
			CroBIM & ConvNeXt-B & BERT & \textcolor{blue}{\textbf{77.41}} & \textcolor{blue}{\textbf{77.55}} & \textcolor{blue}{\textbf{72.62}} & \textcolor{blue}{\textbf{72.83}} & \textcolor{blue}{\textbf{66.74}} & \textcolor{blue}{\textbf{66.38}} & \textcolor{blue}{\textbf{55.92}} & \textcolor{blue}{\textbf{55.93}} & \textcolor{blue}{\textbf{32.17}} & \textcolor{blue}{\textbf{34.07}} & 69.12 & 73.04 & \textcolor{blue}{\textbf{68.70}} & \textcolor{blue}{\textbf{69.33}}  \\    
			\hline
			CroBIM-U & ConvNeXt-B & BERT & \textcolor{red}{\textbf{77.89}} & \textcolor{red}{\textbf{77.73}} & \textcolor{red}{\textbf{72.75}} & \textcolor{red}{\textbf{73.40}} & \textcolor{red}{\textbf{67.90}} & \textcolor{red}{\textbf{67.05}} & \textcolor{red}{\textbf{57.12}} & \textcolor{red}{\textbf{56.85}} & \textcolor{red}{\textbf{33.33}} & \textcolor{red}{\textbf{35.52}} & 69.12 & 73.04 & \textcolor{red}{\textbf{69.15}} & \textcolor{red}{\textbf{69.62}}  \\
			\hline    
		\end{tabular}
	}
	
	\vspace{1.5em}

	\makeatletter
	\def\@captype{figure}
	\makeatother
	\centering
	\includegraphics[width=1\linewidth]{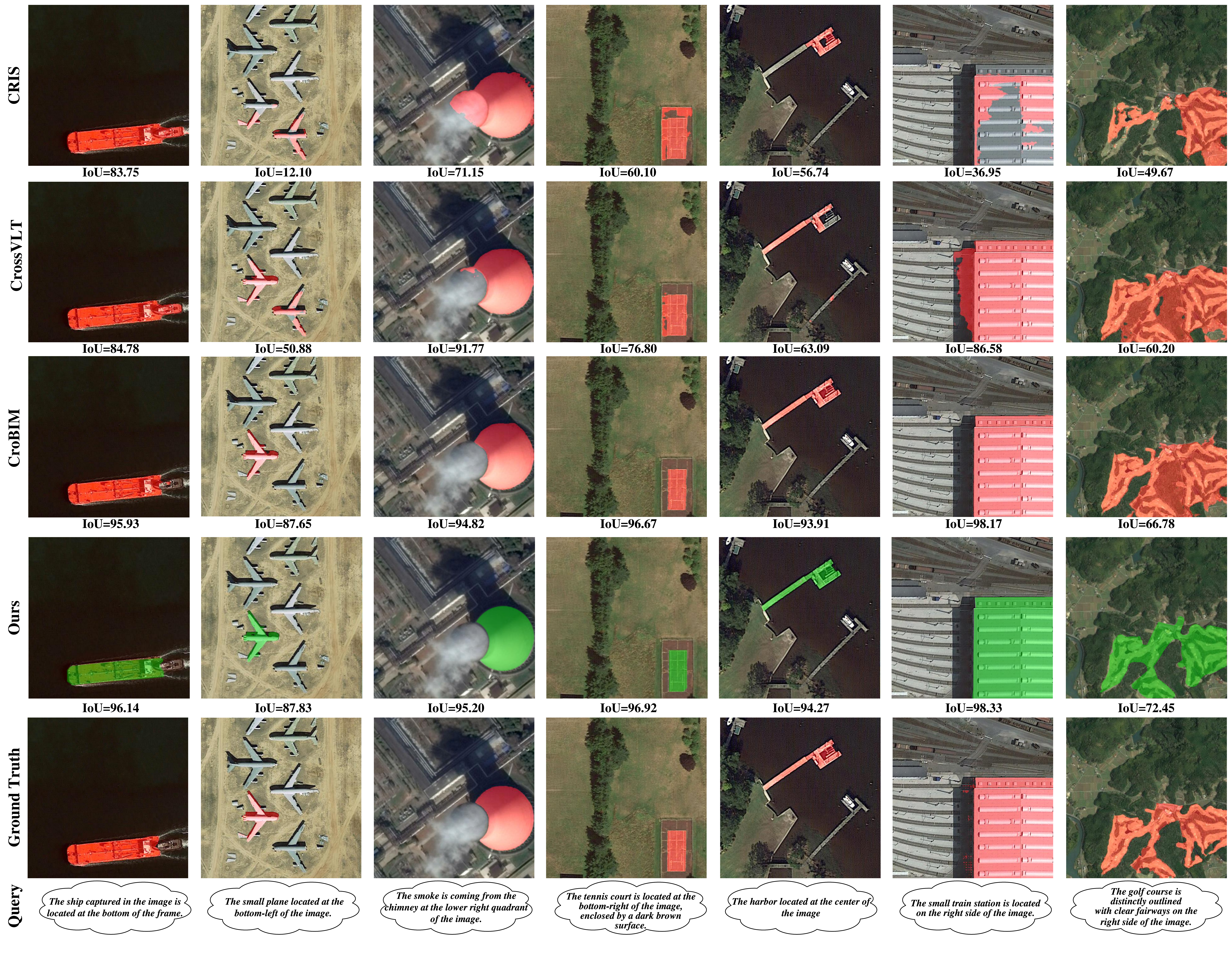}
	\caption{Visualization of segmentation results for CroBIM and comparison methods on the RisBench dataset test set, with corresponding IoU scores displayed.}
	\label{vis_risbench}
\end{table*}

\section{Ablation Study}
\label{section:Ablation}
\begin{figure*}[htbp]
	\begin{center}
		\centerline{\includegraphics[width=1\linewidth]{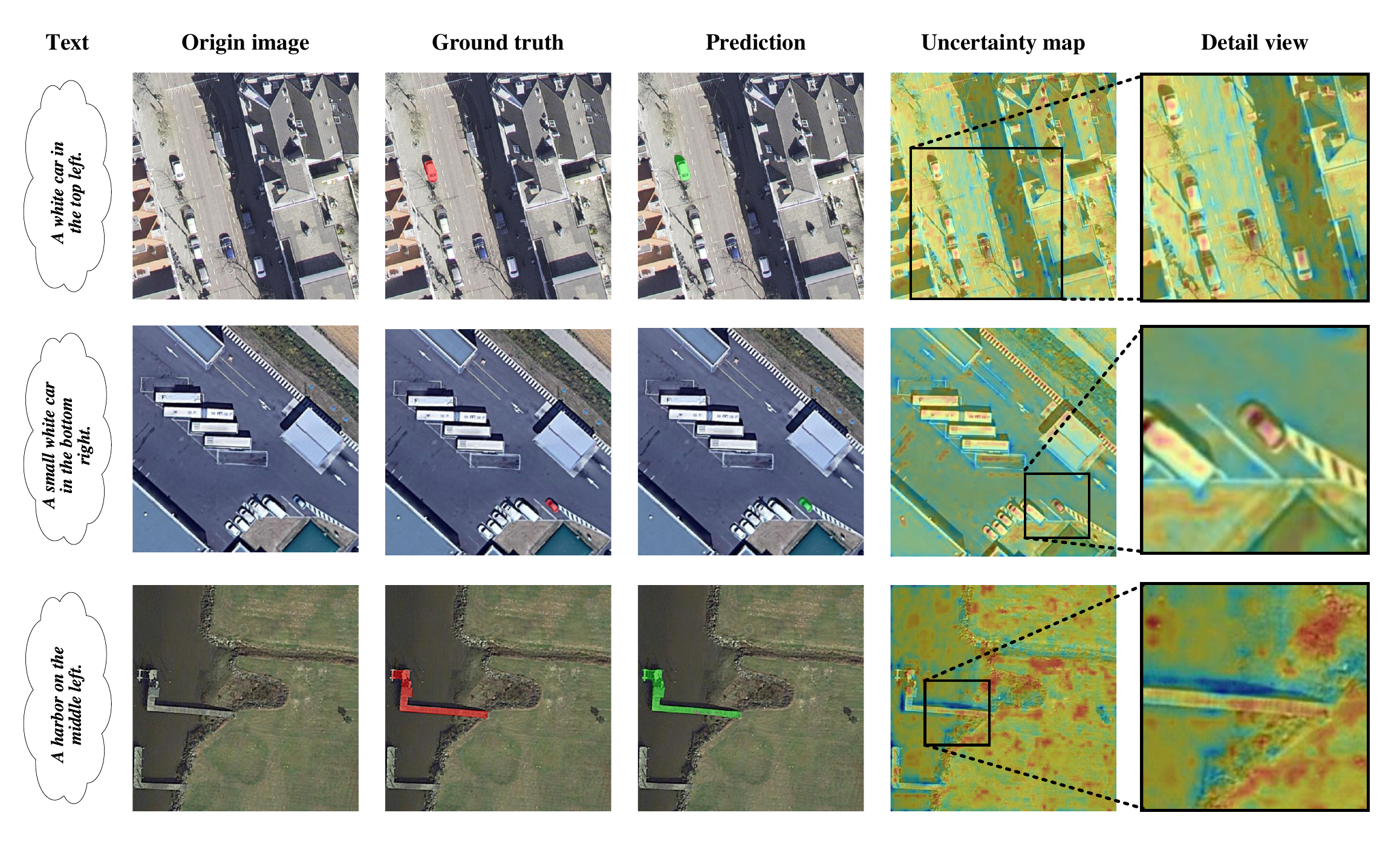}}
		\caption{Qualitative visualization of the uncertainty prior predicted by RUS. From left to right: the referring expression, original image, ground-truth mask (red), model prediction (green), the predicted uncertainty map $U^{p}$ (warmer colors indicate higher uncertainty), and a zoomed-in detail view of the highlighted region. The uncertainty responses concentrate on hard regions such as small/low-contrast targets, confusable surroundings, and boundary/thin-structure areas, indicating that RUS captures an error-prone spatial risk prior for subsequent uncertainty-guided fusion and local refinement. }\label{RUS_ab}
	\end{center}
\end{figure*}
To assess the effectiveness of our designs within the CroBIM framework, we conduct comprehensive ablation studies on the RISBench dataset and report the results on the test split. Unless otherwise specified, all variants share the same backbone, training schedule, and evaluation protocol, and differ only in whether the proposed modules are enabled. We start from the base model (CroBIM) and progressively activate the uncertainty-driven components to form: (i) \textbf{Base} (CroBIM), (ii) \textbf{+UGF }, (iii) \textbf{+UDLR}, and (iv) \textbf{Full} (CroBIM-U). The quantitative comparison is summarized in Table~\ref{ab_risbench}, and we further provide qualitative analysis to interpret how the learned uncertainty prior guides selective fusion and focused local correction.

\begin{table}[tbp]
	\centering
	\caption{Ablation results on RISBench (Test split).}
	\label{ab_risbench}
	\renewcommand{\arraystretch}{1.5}
	\begin{tabular}{l|c|c|c|c|c}
		\toprule
		Setting & Pr@0.5 & Pr@0.7 & Pr@0.9 & oIoU & mIoU \\
		\midrule
		Base                & 77.55 & 66.38 & 34.07 & 73.04 & 69.33 \\
		+UGF                & 77.68 & 66.72 & 34.45 & 73.03 & 69.44 \\
		+UDLR              & 77.60 & 66.93 & 35.10 & 73.02 & 69.53 \\
		Full      & \textbf{77.73} & \textbf{67.05} & \textbf{35.52} & 73.04 & \textbf{69.62} \\
		\bottomrule
	\end{tabular}
\end{table}

\subsection{Effectiveness of RUS}
\label{sec:ab_rus_vis}

We first analyze the Referring Uncertainty Scorer (RUS) qualitatively, since it provides the key intermediate representation---the pixel-wise uncertainty map $U$---that subsequently guides both Uncertainty-Gated Fusion (UGF) and Uncertainty-Driven Local Refinement (UDLR). As described in Sec.~\ref{sec:rus}, $U$ is learned with an online error-aligned supervision signal, encouraging it to respond to locations that are prone to referential confusion or local segmentation errors.

As illustrated in Fig.~\ref{RUS_ab}, each row presents a representative case with the referring expression, original image, ground truth, prediction, the predicted uncertainty map, and a zoomed-in detail view. The three rows correspond to different sources of uncertainty in remote-sensing referring segmentation. \textbf{(Row 1) Dense scenes with similar instances:} the uncertainty map shows strong responses around multiple visually similar candidates, reflecting increased referential ambiguity when targets are surrounded by distractors. \textbf{(Row 2) Small-scale targets:} the uncertainty becomes higher near the referred small object (e.g., a car) due to limited visual evidence and weak saliency, making localization and segmentation more fragile. \textbf{(Row 3) Boundary- and structure-sensitive regions:} for elongated targets with thin boundaries (e.g., the harbor), uncertainty concentrates along edges and local transition areas, where slight geometric deviations can cause boundary shifts, breakage, or adhesion. These observations indicate that $U$ captures a spatial \emph{risk} prior aligned with different failure modes, rather than a generic saliency cue, and thus provides reliable guidance for uncertainty-gated fusion and uncertainty-driven local refinement.

\subsection{Effectiveness of UGF}
\label{sec:ab_ugf}

We then evaluate the contribution of UGF, which uses the uncertainty map to modulate the strength of language injection. UGF is designed to address the dual effect of language cues in remote-sensing scenes: while textual constraints are essential for disambiguation among visually similar instances, uniformly injecting language information can introduce unnecessary perturbations in already confident regions and may even cause referring drift. By gating fusion strength with $U$, UGF strengthens cross-modal constraints in high-uncertainty regions and suppresses ineffective perturbations in low-uncertainty regions.

The ablation results in Table~\ref{ab_risbench} show that enabling \textbf{RUS+UGF} consistently improves segmentation performance over the base model, with the most noticeable gains appearing in \textbf{Pr@0.5} (and typically Pr@0.7). This trend aligns with the intended role of UGF: it primarily improves \emph{instance-level grounding reliability} by reducing confusion among candidate instances under complex descriptions, thereby increasing the probability that the predicted mask overlaps with the correct target at moderate thresholds. Meanwhile, the improvement on Pr@0.9 is relatively smaller, suggesting that UGF mainly affects correct localization and global alignment rather than directly refining fine boundaries.

\subsection{Effectiveness of UDLR}
\label{sec:ab_udlr}

Finally, we assess UDLR, which performs lightweight residual refinement guided by an uncertainty-derived soft region mask. Remote-sensing targets frequently contain thin, fragmented, or low-contrast structures, and the dominant errors tend to cluster around boundaries and fine details, manifesting as merging/bleeding, discontinuities, and boundary shifts \cite{yuan2023rrsis}. UDLR explicitly addresses these issues by focusing corrections on high-risk locations indicated by $U$ and avoiding unnecessary updates in confident regions.

As reported in Table~\ref{ab_risbench}, enabling \textbf{RUS+UDLR} yields consistent improvements over the base model, with larger gains on \textbf{Pr@0.9} and \textbf{mIoU}. This behavior matches the design objective of UDLR: high-threshold precision is especially sensitive to boundary alignment and shape fidelity, and mIoU better reflects quality improvements on diverse targets (including smaller ones) that are common in remote-sensing imagery. In contrast, the change in oIoU is typically marginal, which is expected because oIoU is more influenced by large-area regions, whereas UDLR focuses refinement on localized boundary and thin-structure areas.

\subsection{Complementarity of UGF and UDLR}
\label{sec:ab_comp}

We further combine UGF and UDLR to form the full risk-guided fusion--local correction pipeline. The full model achieves the best overall performance, demonstrating that UGF and UDLR are complementary: UGF improves cross-modal disambiguation and stabilizes global grounding, while UDLR enhances geometric fidelity by correcting boundary and thin-structure errors. Together, they validate the effectiveness of using an error-aligned uncertainty prior to guide both selective fusion and focused local refinement in remote-sensing referring segmentation.

\section{Conclusion}

In this work, we proposed a uncertainty-driven framework for remote-sensing referring segmentation, motivated by the spatially non-uniform reliability of cross-modal grounding and pixel-wise prediction in complex remote-sensing scenes. We introduced a pixel-wise cross-modal referential uncertainty map as a unified spatial prior, and developed an online error-aligned risk learning strategy to make the predicted uncertainty interpretable and consistent with actual model errors. Building on this prior, we designed three plug-and-play modules—RUS, UGF, and UDLR—to enable uncertainty-guided language modulation and focused local correction, strengthening disambiguation in high-risk regions while preserving stable predictions in confident areas.

Extensive experiments on three remote-sensing benchmarks demonstrate that our method consistently improves segmentation accuracy and boundary/detail quality, achieving state-of-the-art performance without modifying the backbone architecture. In future work, we will explore more fine-grained uncertainty modeling and extend the proposed risk-aware paradigm to broader vision--language geospatial understanding tasks.
\ifCLASSOPTIONcaptionsoff
\newpage
\fi
\bibliography{refs}
\vspace{-10 mm} 
\begin{IEEEbiography}[{\includegraphics[width=1in,height=1.25in,clip,keepaspectratio]{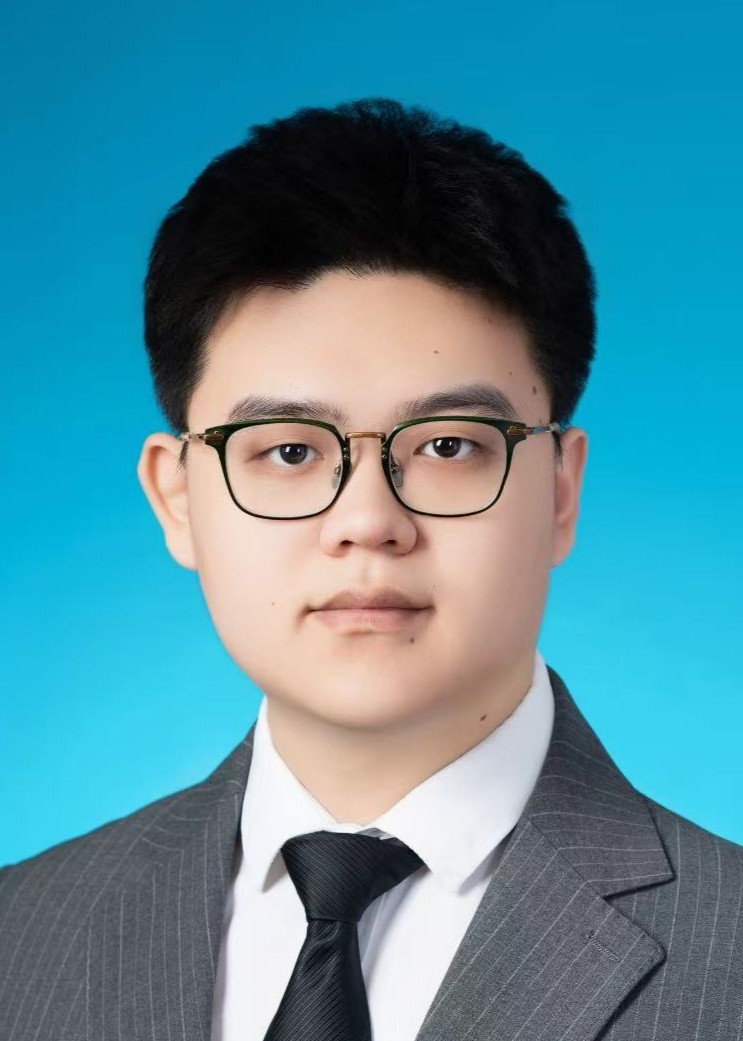}}]{Yuzhe Sun} received his bachelor's degree in Remote Sensing Science and Technology from Harbin Institute of Technology, China, and is currently pursuing the Ph.D. degree in Information and Communication Engineering. His research interests include the development of cross-modal remote sensing models for image and text, downstream tasks of remote sensing images, and the construction of foundational models for remote sensing.
\end{IEEEbiography}
\vspace{-10 mm} 
\begin{IEEEbiography}[{\includegraphics[width=1in,height=1.25in,clip,keepaspectratio]{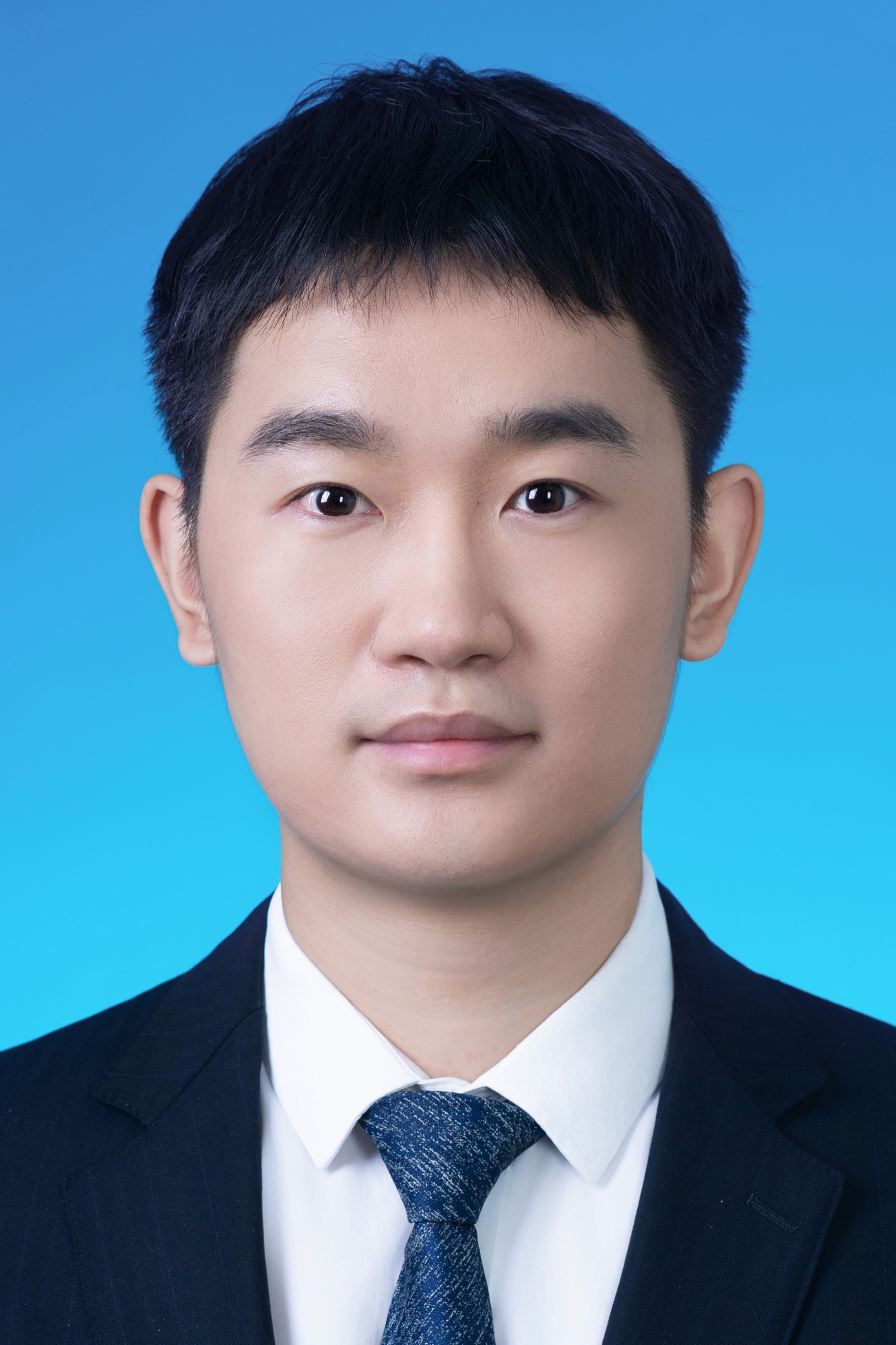}}]{Zhe Dong} received the M.S. degree in software engineering from Harbin Engineering University, Harbin, China, in 2022. He is currently pursuing the Ph.D. degree with the School of Electronics and Information Engineering, Harbin Institute of Technology, Harbin. His research interests are related to the semantic segmentation and self-supervised learning of remote sensing images. 
\end{IEEEbiography}
\vspace{-10 mm} 
\begin{IEEEbiography}[{\includegraphics[width=1in,height=1.25in,clip,keepaspectratio]{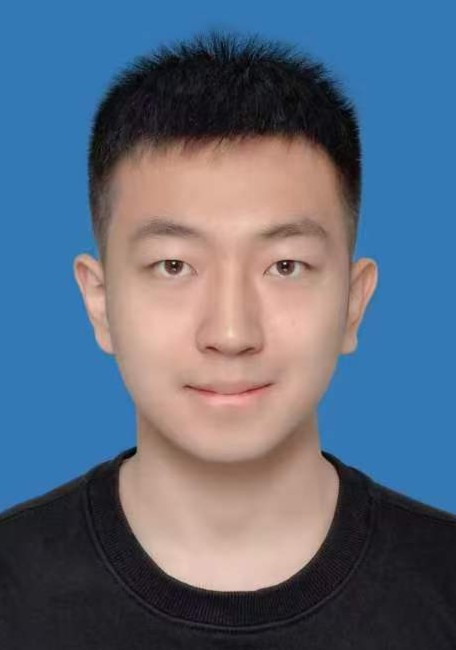}}]{Haochen Jiang}  received his bachelor's degree in Electronic Information Science and Technology from Jilin University, China, and is currently pursuing a Ph.D. degree in Information and Communication Engineering. His research interests include multimodal learning with remote sensing imagery and text, downstream tasks for remote sensing image understanding, and the development of foundation models in remote sensing. 
\end{IEEEbiography}
\vspace{-10 mm} 
\begin{IEEEbiography}[{\includegraphics[width=1in,height=1.25in,clip,keepaspectratio]{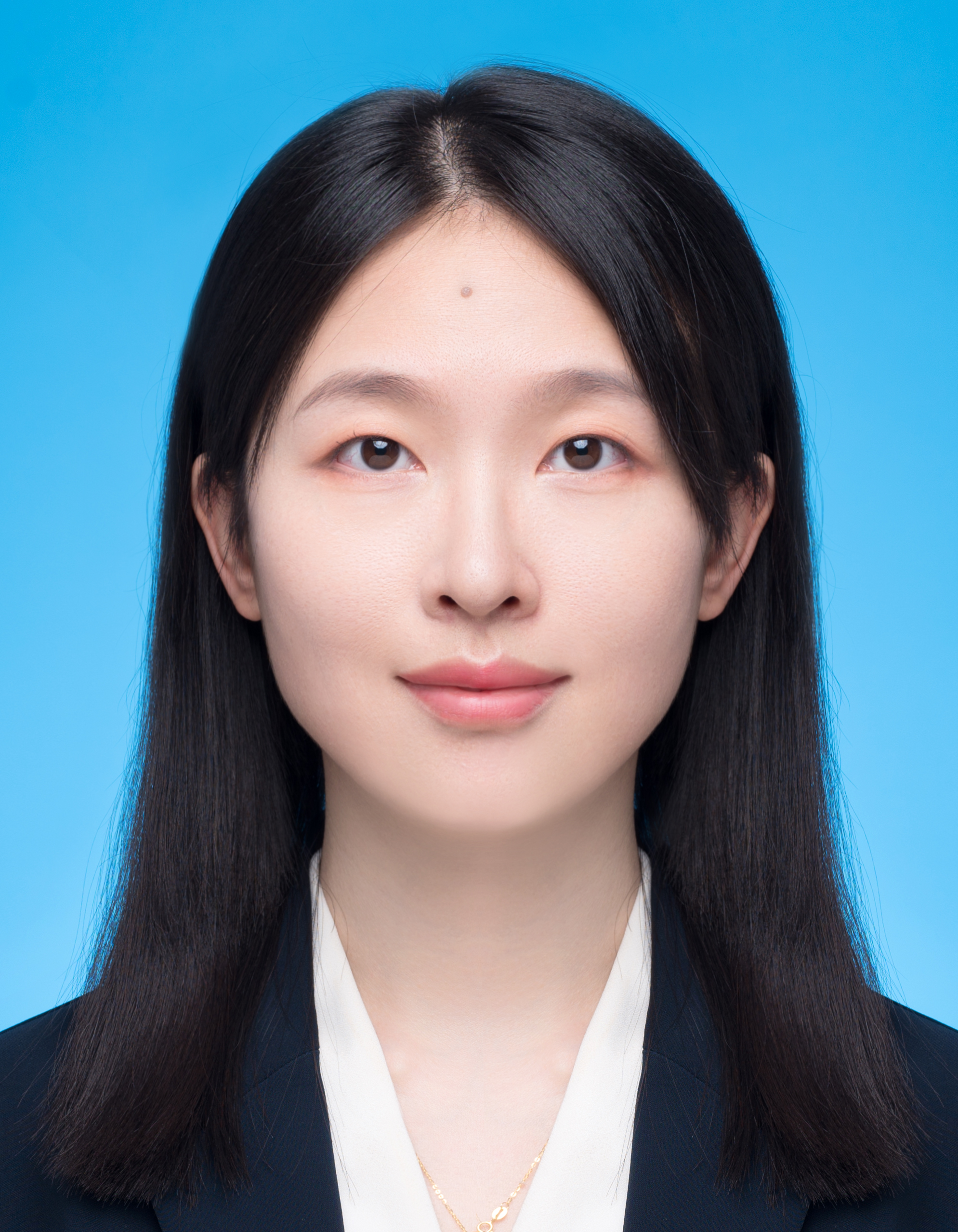}}]{Tianzhu Liu} (Member, IEEE) received the Ph.D. degree in information and communication engineering from the Harbin Institute of Technology (HIT), Harbin, China, in 2019. She was a Lecturer with the School of Electronics and Information Engineering, HIT, where she is currently an Associate Professor and a Post-Doctoral Fellow. Her research interests include image processing in hyperspectral remote sensing, especially multimodal hyperspectral image classification. Dr. Liu won the Third Prize in Student Paper Contest at the International Geoscience and Remote Sensing Symposium (IGARSS) in 2019. She serves as an Associate Editor for the \textit{IEEE Journal of Selected Topics in Applied Earth Observations and Remote Sensing} and a Reviewer for several international journals, such as the \textit{IEEE Transactions on Geoscience and Remote Sensing} and \textit{Neurocomputing}.
\end{IEEEbiography}
\vspace{-10 mm} 
\begin{IEEEbiography}[{\includegraphics[width=1in,height=1.25in,clip,keepaspectratio]{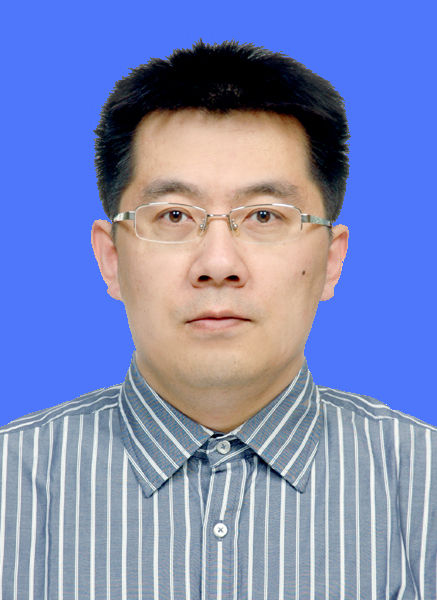}}]{Yanfeng Gu} (M’06-SM’16) received the Ph.D. degree in information and communication engineering from Harbin Institute of Technology, Harbin, China, in 2005. He joined as a Lecture with the School of Electronics and Information Engineering, Harbin Institute of Technology (HIT). He was appointed as Associate Professor at the same institute in 2006; meanwhile, he was enrolled in first Outstanding Young Teacher Training Program of HIT. From 2011 to 2012, he was a Visiting Scholar with the Department of Electrical Engineering and Computer Science, University of California, Berkeley, CA, USA. He is currently a Professor with the Department of Information Engineering, HIT, Harbin, China. He has published more than 100 peer-reviewed papers, four book chapters, and he is the inventor or coinventor of 20 patents. His research interests include space intelligent remote sensing and information processing, multimodal hyperspectral remote sensing, spaceborne time-series image processing.
\end{IEEEbiography}
\end{document}